%% file: sn-article.tex
\expandafter\def\csname ver@program.sty\endcsname{3000/12/31}

\RequirePackage{tikz}

\documentclass[default,iicol]{sn-jnl}

\usepackage{amsmath,amssymb,amsfonts}
\usepackage{graphicx}
\usepackage{textcomp}
\usepackage{tabularx}
\usepackage{float}
\usepackage{url}
\usepackage[acronym]{glossaries}
\usepackage{cleveref}
\usepackage{pgfplots}
\usetikzlibrary{arrows.meta, positioning, quotes}

\def\BibTeX{{\rm B\kern-.05em{\sc i\kern-.025em b}\kern-.08em
    T\kern-.1667em\lower.7ex\hbox{E}\kern-.125emX}}

\DeclareMathOperator*{\argmax}{arg\,max}

\usepackage{tikz-qtree}
\usepackage{hyperref}

\jyear{2023}%

\theoremstyle{thmstyleone}%
\theoremstyle{thmstyletwo}%

\theoremstyle{thmstylethree}%

\begin{document}

\newacronym{MCTS}{MCTS}{Monte Carlo Tree Search}
\newacronym{RL}{RL}{Reinforcement Learning}
\newacronym{UCT}{UCT}{Upper Confidence Bound for Trees}
\newacronym{PUCT}{PUCT}{Predictor + UCT}
\newacronym{MSE}{MSE}{mean squared error}
\newacronym{SMILES}{SMILES}{simplified molecular-input line-entry system}
\newacronym{TSP}{TSP}{Traveling Salesman Problem}
\newacronym{CNN}{CNN}{convolutional neural network}
\newacronym{MDP}{MDP}{Markov decision process}

\title[Article Title]{Beyond Games: A Systematic Review of Neural Monte Carlo Tree Search Applications}


\author*[1]{\fnm{Marco} \sur{Kemmerling} }\email{marco.kemmerling@ima.rwth-aachen.de} 

\author[1]{\fnm{Daniel} \sur{L{\"u}tticke}}

\author[1]{\fnm{Robert H.} \sur{Schmitt}}

\affil[1]{\orgdiv{Information Management in Mechanical Engineering (WZL-MQ/IMA)}, \\ \orgname{RWTH Aachen University}, \orgaddress{\city{Aachen}, \country{Germany}}}

\abstract{The advent of AlphaGo and its successors marked the beginning of a new paradigm in playing games using artificial intelligence. This was achieved by combining Monte Carlo tree search, a planning procedure, and deep learning. While the impact on the domain of games has been undeniable, it is less clear how useful similar approaches are in applications beyond games and how they need to be adapted from the original methodology. We review 129 peer-reviewed articles detailing the application of neural Monte Carlo tree search methods in domains other than games. Our goal is to systematically assess how such methods are structured in practice and if their success can be extended to other domains. We find applications in a variety of domains, many distinct ways of guiding the tree search using learned policy and value functions, and various training methods. Our review maps the current landscape of algorithms in the family of neural monte carlo tree search as they are applied to practical problems, which is a first step towards a more principled way of designing such algorithms for specific problems and their requirements.  }




\maketitle

\section{Introduction}
The combination of \gls{MCTS} and deep learning led to the historical event of the computer program AlphaGo beating a human champion in the game of Go \cite{silver_mastering_2016}, which had been considered beyond the capabilities of computational approaches for a long time. Since then, such approaches, which we term \emph{neural \gls{MCTS}} in this review, have enjoyed a huge amount of popularity. They have been applied to many other games and yield promising results in the field of general game playing \cite{schrittwieser_mastering_2020, silver_general_2018}. 

While the effectiveness of neural \gls{MCTS} in game contexts has been clearly established, the transfer of such approaches to non-game-playing applications is still a fairly recent development and hence less well understood. 

Some surveys and reviews of \gls{MCTS} approaches have been published in the past, e.g. Mandziuk \cite{mandziuk_mctsuct_2018} provides a survey of selected \gls{MCTS} applications on non-game-playing problems. However, it only examines two different applications, neither of which feature any neural guidance. 

A more extensive survey is provided in \cite{swiechowski_monte_2022}, which also features a brief section on the combination of \gls{MCTS} and deep learning. However, their focus is on games rather than other applications. 

We are not aware of any reasonably extensive review of applications of neural \gls{MCTS} methods in non-game-playing domains. We believe that such a review can shed light on the extent to which such methods can be transferred to more practical problems and how neural \gls{MCTS} methods can be designed to cope with the requirements of different use cases. To this end, we hope to provide a contribution towards a more principled guide towards the design of neural \gls{MCTS} approaches. 

The following research questions guide our review: 
\begin{enumerate}
    \item In which disciplines, domains, and application areas is neural \gls{MCTS} used? What are commonalities and differences in the observed applications?
    \item What differences in the design of neural \gls{MCTS} methods can be observed compared to applications in games?
    \item Where and how can neural guidance be used during the tree search? 
\end{enumerate}

To address these questions, we perform a systematic literature review and analyze the resulting literature. Starting with a keyword search in multiple databases, we filter articles for relevance, perform additional forward and backward searches, and extract a set of predefined data items from each article included in the review. The detailed review process is described in \cref{sec:research_methodology}. Before, we provide a brief introduction to the concepts of reinforcement learning, \gls{MCTS}, and AlphaZero in the next section. The remaining sections begin with a focus on the problems described in the surveyed publications in \cref{sec:applications}, and continue with an examination of the employed methods in \cref{sec:results_nmcts_methodologies}. We end our review with a brief discussion in \cref{sec:discussion_conclusion}. 


\section{Reinforcement Learning \& Neural MCTS} 
\label{sec:background}
\subsection{Reinforcement Learning}
\gls{RL} is a paradigm of machine learning, in which \emph{agents} learn from experience collected from an \emph{environment}. To do so, an agent observes the \emph{state} $s$ of the environment and executes an \emph{action} $a$ based on this state. Upon acting, agents receive a \emph{reward} $r$ and observe a new state $s'$. A problem which follows this kind of formulation is called a \gls{MDP} if the new state $s'$ only depends on the state $s$ immediately preceding it and the action $a$ of the agent. The agent's goal in such an \gls{MDP} is to maximize the return, i.e. the expected long-term cumulative reward, by learning an appropriate \emph{policy} $\pi$, i.e. a behavioural strategy that prescribes an action, or a probability distribution over actions, for a given state \cite{sutton_reinforcement_2018}. 

Such a policy can be learned directly from experience, e.g. through policy gradient methods, or it can be derived from a learned action-value function. An action-value function $Q^\pi(s,a)$ estimates the value, i.e. the expected return, of taking action $a$ in state $s$. From such a learned action-value function, a deterministic policy can be derived by greedily choosing the best action, while a stochastic policy can be derived by sampling actions proportionally to their value. In addition to the policy- and value-based approaches described so far, hybrid approaches which synergistically learn both policy and value functions are often employed as well. Such methods are called actor-critic approaches and often learn the state-value function $V^\pi(s)$ instead of the action-value function $Q^\pi(s,a)$. The former computes the expected return of state $s$ when following policy $\pi$, while the latter computes the expected return of state $s$ when first executing action $a$ and following $\pi$ in subsequent steps \cite{sutton_reinforcement_2018}. The superscript $\pi$ is often omitted for more concise notation. 

In contrast to model-free approaches, in model-based \gls{RL}, a \textit{model} of the environment is used for \textit{planning}. A model simulates the dynamics of the environment either exactly or approximately. Planning simply refers to the simulation of experience using a model and planning approaches can be categorized into \emph{background planning} and \emph{decision-time planning}. In the former, the training data consisting of real experiences collected from the environment is augmented with imagined experience generated from a model. In the latter, the action selection at a given time-step is dependent on planning (ahead) using a model, i.e. the consequences of different choices of actions are imagined to improve the policy for the current state \cite{moerland_unifying_2022}. \gls{MCTS}, further explained in the following, can be considered a form of decision-time planning. 

\subsection{The Connection Between RL and MCTS}

\gls{MCTS} arose as a heuristic search method to play combinatorial games by performing a type of tree search based on random sampling. In such games, a player has to decide which action to perform in a given state in order to maximize an outcome $z$ at the terminal state of the game. While \gls{MCTS} has not been traditionally thought of as a type of reinforcement learning, the scenario described here bears strong similarities to the formulation of reinforcement learning problems given earlier and some authors have explored this connection in detail \cite{vodopivec_monte_2017}. To avoid ambiguity, we will not use the term reinforcement learning to refer to \gls{MCTS} in this article. Similarly to \gls{RL}, \gls{MCTS} also produces a policy $\pi_{MCTS}$ and a value estimate $v_{MCTS}$. In \gls{MCTS}, the policy is produced for a given state $s$ by a multi-step look-ahead search, i.e. by considering future scenarios and determining which sequence of actions will lead to favourable outcomes starting from $s$. This policy is produced anew for every encountered state, i.e. the determined policy does not generalize to states other than the one currently encountered. In contrast, traditional \gls{RL} produces policies by learning from past experience that aim to generalize to unseen situations. At decision-time, no forward search is performed and an action is simply chosen based on the policy learned from past experience. In a sense, \gls{MCTS} looks into the future, while traditional \gls{RL} looks back to the past to determine actions. As a consequence, \gls{RL} requires computationally expensive upfront training but incurs negligible computational cost at decision time, while \gls{MCTS} requires no training, but performs computationally expensive planning at decision time. 

\subsection{MCTS}
\label{subsec:mcts}
The general idea of \gls{MCTS} is to iteratively build up a search tree of the solution space by balancing the exploration of infrequently visited tree branches with the exploitation of known, promising tree branches. This is accomplished by the repeated execution of four different phases: selection, expansion, evaluation, and back-propagation. In the selection phase, starting from the root node, actions are chosen until a leaf node $s_L$ is encountered. New children are then added to this leaf node in the expansion phase and their value is estimated in the evaluation phase. Finally, the values of the newly added nodes are back-propagated up the tree to update the values of nodes along the path to $s_L$. In the following, we describe each of these phases in more detail. 

\paragraph{Selection} In the selection phase, starting from the root node, an action is chosen according to some mechanism. This leads to a new state, in which the selection mechanism is applied again. The process is repeated until a leaf node is encountered. 

The mechanism of action selection is referred to as the \emph{tree policy}. While different mechanisms exist, the UCB1 \cite{auer_finite-time_2002} formula is a popular choice. When \gls{MCTS} is used with the UCB1 formula, the resulting algorithm is called  \gls{UCT} \cite{kocsis_bandit_2006}. In \gls{UCT}, the action selection is defined as follows: 
\begin{equation}
    a = \argmax_a \frac{W(s,a)}{N(s,a)} + c \ \sqrt{\frac{ln \ N(s)}{N(s,a)}}
\end{equation}

where $W(s,a)$ represents the number of wins encountered in the search up to this point when choosing action $a$ in state $s$, $N(s,a)$ the number of times $a$ has been selected in $s$, and $N(s)$ the number of times $s$ has been visited. The left part of the sum encourages exploitation of actions known to lead to favourable results where the fraction $\frac{W(s,a)}{N(s,a)}$ can be seen as an approximation of $Q(s,a)$. The right part of the sum encourages exploration by giving a higher weight to actions that have been visited less often compared to the total visit count of the state. Exploration and exploitation are balanced by the exploration constant $c$. For game outcomes $z \in [0,1]$, the optimal choice of $c$ is $c = \frac{1}{\sqrt{2}}$ \cite{kocsis_improved_2006}, but for rewards outside this range, $c$ may have to be adjusted \cite{browne_survey_2012}. 

\paragraph{Expansion} After repeated application of the selection step, the search may arrive at a node with unexpanded potential children. Once this happens, one or more children of the node will be expanded. There are some possible variations in this phase. In some cases, all possible children are expanded when a leaf node (a node with \emph{no} children) is encountered. In other cases, a single child is expanded when an expandable node (a node with \emph{some} as of yet unexpanded children) is encountered. Expanding all children right away may lead to undesirable tree growth depending on the application.  

\begin{figure}
    \centering
    \resizebox{\columnwidth}{!}{
    \input{Fig1}
    }
    \caption{Search tree as seen at a given time during the search. The current node is indicated with a thick border, as are the edges that were traversed in the current iteration of the search. $s_1$ is currently a leaf node and its potential children $s_3$ and $s_4$ are considered for expansion, as their dotted edges signify. The terminal nodes $s_7$, $s_8$, $s_9$, and $s_10$ are represented by rectangular nodes. }
    \label{fig:mcts_tree_construction}
\end{figure}
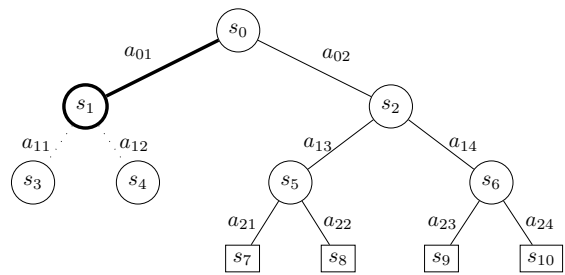

In some literature, expandable nodes are also called leaf nodes. For clarity, we will only use the term leaf node to refer to true leaf nodes without any children in this article. Note that a leaf node is not the same as a terminal node, with the former merely being the current end of a tree branch, while the latter is a node that represents an end state of the game (see \cref{fig:mcts_tree_construction}). 

\paragraph{Evaluation} Once a node has been expanded, it is evaluated to initialize $W(s,a)$ and $N(s,a)$. This evaluation is sometimes also called \emph{simulation}, \emph{roll-out}, or \emph{play-out} and consists of playing the game starting from the newly expanded node until a terminal state is encountered. The outcome $z$ at the terminal state is the result of the evaluation. The game is played according to a \emph{default policy}, which determines the sequence of actions between the newly expanded node and the terminal one. In the simplest case, the default policy samples actions uniformly randomly \cite{kocsis_improved_2006}. 

Instead of evaluating newly expanded nodes, it is also possible to only evaluate leaf nodes and and leave the evaluation of the newly expanded nodes for a later point in the search, when they are again encountered as leaf nodes themselves. 

\paragraph{Back-propagation}
The outcome $z$ from the evaluation phase is propagated up the tree to update $W(s,a)$ among the preceding nodes. The visit counts of all selected nodes are incremented as well. \newline

\noindent Once the back-propagation phase is finished, the process starts anew from the selection phase until a predefined simulated budget is reached. 

\subsection{AlphaZero}
The program known as \textit{AlphaGo} drew attention for being the first computer program to beat a professional human player in a full size game of Go \cite{silver_mastering_2016} by combining deep learning and \gls{MCTS}. While AlphaGo relied on supervised pre-training on human expert moves prior to reinforcement learning, its successor, AlphaGo Zero, was only trained using reinforcement learning by self-play. It further simplified the training by reducing the number of employed neural networks. AlphaGo Zero was still developed specifically for the board game Go and incorporated some game-specific mechanisms. In contrast, the next iteration of the AlphaGo family, AlphaZero, is more generic and can be applied to a variety of board games. 

The algorithms introduced in this subsection are all examples of neural \gls{MCTS}, i.e. \gls{MCTS} guided by neural networks. While we focus on the AlphaZero family here due to its popularity, similar ideas were independently proposed under the name of \emph{Expert Iteration} \cite{anthony_thinking_2017}. In the following, we provide more details on AlphaZero as one representative of neural \gls{MCTS} methods utilized for games. 

Like regular \gls{MCTS}, AlphaZero follows the four phases of selection, expansion, evaluation, and back-propagation. Some of the phases are assisted by a neural network $f_\theta$, which, given a state, produces a policy vector $\mathbf{p}$, i.e. a probability distribution over all actions, and an estimate $v$ of the state value. 

The selection phase in AlphaZero uses a variant of the \gls{PUCT} formula \cite{rosin_multi-armed_2011}:
\begin{equation}
    a = \argmax_a Q(s,a) + c \ P(s,a) \ \frac{\sqrt{N(s)}}{1+N(s,a)}
\end{equation}

where $P(s,a)$ denotes a prior probability of choosing action $a$ in state $s$ given by $f_\theta$ \cite{silver_general_2018}.

Once the selection phase reaches a leaf node $s_L$, it is evaluated by the neural network $(\mathbf{p}, v) = f_\theta(s_L)$.  The leaf node is then fully expanded and its children initialized with $N(s_L, a) = 0, W(s_L, a) = 0, Q(s_L, a) = 0, P(s_L, a) = p_a$. In the back-propagation step, the statistics of each node including and preceding $s_L$ are then updated as: $N(s_t, a_t) = N(s_t, a_t) + 1, W(s_t, a_t) = W(s_t, a_t) + v, Q(s_t, a_t) = \frac{W(s_t, a_t)}{N(s_t, a_t)}$ \cite{silver_general_2018}. 

Note that the expansion and evaluation phases are interwoven here to some degree and do not strictly follow the order of the phases in standard \gls{MCTS}. In this review, we are generally not overly concerned with the \gls{MCTS} phases as a strictly ordered set of algorithmic steps, but more with the function each phase fulfills in the tree search.

Once the search budget is exhausted, an improved policy $\pi_{MCTS}$ is derived from the visit counts $N(s,a)$ in the tree and a corresponding value estimate $v_{MCTS}$ is extracted. The produced policy $\pi_{MCTS}$ and value estimate $v_{MCTS}$ are then used as training targets to further improve $f_\theta$. Since AlphaZero plays two-player games, some mechanism is required to determine the actions of the second player. In what is called \emph{self-play}, the actions for the second player are chosen by (some version of) the same policy currently being trained for the first player \cite{silver_general_2018}.

As a model-based \gls{RL} algorithm, AlphaZero needs a model of the environment to perform the tree search. This model is simply assumed to be given, although further extensions such as MuZero \cite{schrittwieser_mastering_2020} demonstrate that such a model can be learned from collected experience during the search. 

To recapitulate, the neural guidance in AlphaZero, consists of evaluating nodes by using $f_\theta$ to compute $v$ and $\mathbf{p}$, which are then used in the selection phase. The way neural guidance is used in AlphaZero is not the only possible form of neural guidance. Other possibilities to guide the search exist, as will become apparent in \cref{sec:results_nmcts_methodologies}.

\section{Research Methodology}
\label{sec:research_methodology}
We perform a systematic sequential literature review as outlined in \cite{vom_brocke_standing_2015}, i.e. we follow a series of pre-defined steps in a given sequence consisting of a keyword search in multiple databases, a screening process to filter for relevant articles, a forward and backward search, data extraction from all included articles, followed by analysis and synthesis of the results. 

While we aim to take a neutral position and hence do not want to limit the collected literature on arbitrary grounds, a comprehensive literature search attempting to capture \emph{all} the relevant literature is infeasible due to incurred time-requirements. Instead, we aim to balance feasibility and coverage by collecting a representative sample of the existing literature by limiting ourselves to a keyword search with a defined set of keywords in a limited number of databases. Both the set of keywords as well as the set of databases could be enlarged to arrive at more comprehensive results. 

\subsection{Search Query \& Databases}
To find relevant publications, we derive three types of keywords: 

\begin{figure*} 
\centering
\input{Fig2}
\caption{The literature search and screening process starting from a set of keywords to the final set of publications to be included in the review.}
\label{fig:screening_process}
\end{figure*}
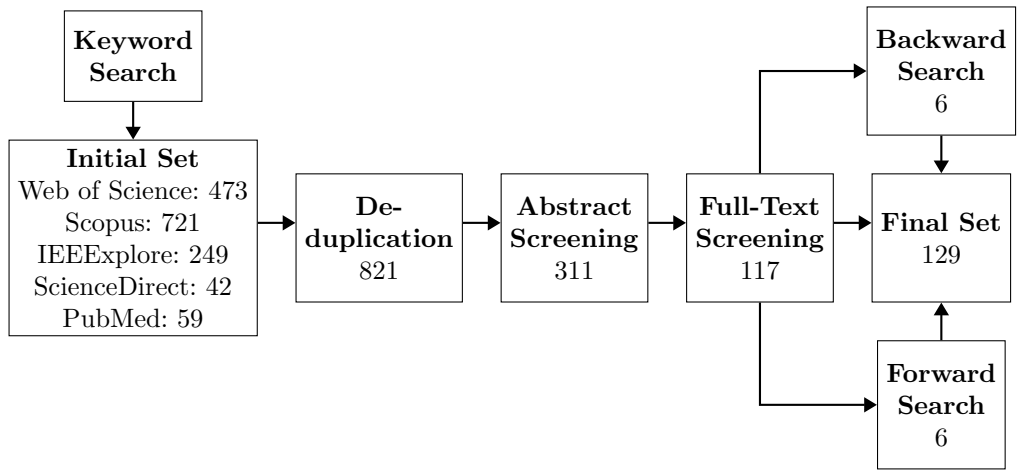

\begin{enumerate}
    \item  Based on neural \gls{MCTS} being a combination of \gls{MCTS} and neural networks or traditional reinforcement learning:
    \begin{itemize}
        \item ``reinforcement learning'' AND ``monte carlo tree search''
        \item ``neural monte carlo tree search''
        \item ``neural MCTS'' 
    \end{itemize}
    \item  Based on \gls{MCTS} providing the ability to perform decision-time planning in a model-based reinforcement learning setting:
    \begin{itemize}
        \item ``decision-time planning'' AND ``reinforcement learning''
    \end{itemize}
    \item  Based on the names given to algorithms in the AlphaGo family:
    \begin{itemize}
        \item AlphaGo
        \item AlphaZero
        \item MuZero
    \end{itemize}
\end{enumerate}

Each of the partial search strings expressed after a bullet point above is connected with an OR operator to arrive at one overall search query. 

We use this query to search for publications in the databases Web of Science, IEEExplore, Scopus, ScienceDirect, and PubMed. The search query is applied to the abstract, title, and keywords.

\subsection{Eligibility Criteria \& Screening}
To be included in the review, a given publication must fulfill a predefined set of eligibility criteria:

\begin{enumerate}
    \item \textbf{Must feature an application of \gls{MCTS} guided by a neural network.} This excludes publications which are purely reviews or surveys. By guidance, we mean that a learned policy or value function is used in at least one of the phases of the tree search. These functions can of course be learned by other means than neural networks. We explicitly only consider neural network based approaches here because preliminary searches showed that attempting to include other methods leads to many more irrelevant search results while providing relatively little additional value. 
    \item \textbf{The publication must contain at least some amount of validation of the presented approach.} Purely conceptual articles are not considered. 
    \item \textbf{The problem to which neural \gls{MCTS} is applied must not be a game.}  While many impressive results have been achieved using neural \gls{MCTS} in game playing, we are interested in determining whether such approaches transfer to other applications as well. We do consider applications that are not typically considered a game, but have been modelled as a game to facilitate the use of neural \gls{MCTS}. 
    \item \textbf{Publication language must be English.\\} 
\end{enumerate}

For each of the publications retrieved during the keyword search described in the previous section, we assess its eligibility according to the above criteria. After the removal of duplicates, the screening process is conducted in two phases. In the first phase, we only examine the abstract of each article and discard it if it is clear that at least one of the above criteria is not fulfilled. If there is any ambiguity, we reexamine the article in a second phase, where we repeat the process using the full text of the article. Any article that is not discarded in this second phase will be included in our review. 

\subsection{Forward and Backward Search}
After the abstract and full-text screening described above, we perform a forward and backward search \cite{webster_analyzing_2002} based on the set of publications which has passed the screening process. That is, for every article, we check its references for further relevant publications and also look for publications that in turn reference the articles which passed our screening. 

In this step, we notice that among the additionally identified literature, many simply cite an approach already included in our review for a similar application without performing any modifications or providing additional details. We do not include such publications in our review since they do not provide any benefit in addressing our research questions. The eligibility criteria described above apply to the results from forward and backward search as well. 

The full search process from keyword search to our final set of publications is visualized in \cref{fig:screening_process}.

\subsection{Data Extraction}

\begin{table*}[h]
\begin{center}
\caption{Information extracted from each article after screening.}\label{tab:dataextraction}%
\begin{tabular}{@{}p{0.15\textwidth}p{0.2\textwidth}p{0.25\textwidth}p{0.25\textwidth}@{}}

\toprule
Application & Neural \gls{MCTS} & Implementation & Other \\ 
\midrule
Problem & Selection Details & Network Type &  Country \\
Time & Expansion Details & Activation Functions & Organization Type \\
Reward Scale & Evaluation Details & Policy Loss & Comparison \gls{MCTS} \\
Horizon & Use of Self-play & Value Loss & Comparison Model-free \\
Transitions & Training Method & MCTS Hyperparameters & Comparison Baselines \\
Action space & Learned Model & Code Availability & Used Hardware \\
\botrule
\end{tabular}
\end{center}
\end{table*}   
For every article resulting from the search process described above, we extract information along a set of predefined categories. These are given in \cref{tab:dataextraction}, where \emph{problem} refers to a short description of the examined problem, \emph{time} is either continuous or discrete, \emph{horizon} either finite or infinite, \emph{transitions} either deterministic or stochastic. 

Since it can be difficult to extract information from works originating from different disciplines with differing terminology and varying descriptions of details, we generally err on the side of providing incomplete rather than wrong information. 

Some of the collected information did not lead to notable insights and will hence not be discussed in this review. This includes the activation functions, the author affiliations, and comparisons of the neural \gls{MCTS} approach with model-free \gls{RL} or \gls{MCTS} without neural guidance. In such comparisons, neural \gls{MCTS} tends to find solutions with superior quality, but this could simply be due to positive-results publication bias.

\section{Neural MCTS Applications}
\label{sec:applications}
\subsection{Application Fields}
To determine the applicability of neural \gls{MCTS} outside of game-playing, we survey the areas of application to which neural \gls{MCTS} has been transferred. The algorithmic details of individual approaches are analyzed in \cref{sec:results_nmcts_methodologies}. Here, we simply outline where such approaches are applied. We find applications in a wide variety of domains including chemistry, medicine, production, electrical engineering, and computer science. In the following, we assign each publication to a specific application area. Note that his merely serves the purpose of creating an overview of the research landscape. Many articles could be assigned to more than one category, and the choice of categories itself could have been made in many different ways. 

\paragraph{Chemistry} In the chemical literature in particular, neural \gls{MCTS} has received considerable attention. It has been used to perform synthesis planning, de novo molecular design, protein folding, and more (see \cref{tab:chemistry_applications}). In many cases, states are represented by a \gls{SMILES} string, a notation allowing for the representation of molecular structure \cite{weininger_smiles_1988}. Such a string can then be iteratively constructed during the \gls{MCTS} process until a viable molecule is found or the attempt is discarded. 

Molecular applications appear to be a  comparatively mature branch of neural \gls{MCTS} research, as evidenced by the fact that authors are building on each others work and by the existence of standardized, commonly used implementations such as ChemTS \cite{yang_chemts_2017}. This is an exception rather than the norm, as most literature is more disjointed and most other works either do not specify any implementation or use a custom one (see appendix \ref{sec:app_implementations} for a detailed list of observed implementations). 

\begin{table}[h]
\begin{center}
\caption{Applications in chemistry.}\label{tab:chemistry_applications}%
\begin{tabular}{@{}p{0.1\columnwidth}p{0.75\columnwidth}@{}}

\toprule
Source & Application \\ 
\midrule
\cite{deng_neural-augmented_2022} & Protein Folding \\ 
\cite{erikawa_mermaid_2021} & Lead Generation (Drug Discovery) \\ 
\cite{mao_learning_2021} & RNA folding \\ 
\cite{sridharan_deep_2022} & Reconstructing Molecular Structure from Nuclear Magnetic Resonance Spectra \\ 
\cite{sv_multi-objective_2022} & Optimization of De Novo Stable Organic Radicals for Aqueous Redox Flow Batteries \\ 
\midrule
\cite{genheden_aizynthfinder_2020} & \multirow{7}{*}{Synthesis Planning} \\ 
\cite{ishida_ai-driven_2022} &  \\ 
\cite{segler_planning_2018} &  \\ 
\cite{segler_towards_2017} &  \\ 
\cite{thakkar_datasets_2020} & \\ 
\cite{wang_towards_2020} &  \\ 
\cite{zhang_chemistry-informed_2022} & \\ 
\midrule
\cite{li_structure-based_2021} & \multirow{6}{*}{De Novo Molecule Design} \\ 
\cite{ma_structure-based_2021} &  \\ 
\cite{motomichi_novo_2021} &   \\ 
\cite{qian_alphadrug_2022} &  \\ 
\cite{srinivasan_artificial_2021} & \\ 
\cite{yang_chemts_2017} &  \\ 

\botrule
\end{tabular}
\end{center}
\end{table}   

\paragraph{Material Science} Closely related is the domain of material science (see \cref{tab:material_applications}). In some cases, the \gls{SMILES} representation is used here as well, such as in the design of metal-organic frameworks \cite{tang_design_2021, xiangyu_machine_2020}, where metal ions and organic ligands are combined to create structures of various shapes. 

In other cases, neural \gls{MCTS} is used to optimize the thickness of alternating layers of two materials in a multilayer structure such that certain desired properties are achieved \cite{dieb_optimization_2020} and to generate models which describe the mechanical behaviour of materials in various circumstances \cite{wang_meta-modeling_2019}. 
\begin{table}[h]
\begin{center}
\caption{Applications in material science.}\label{tab:material_applications}%
\begin{tabular}{@{}p{0.1\columnwidth}p{0.75\columnwidth}@{}}

\toprule
Source & Application \\ 
\midrule
\cite{dieb_optimization_2020} & Depth-graded Multilayer Structure Design for X-ray Optics \\ 
\cite{wang_meta-modeling_2019} & Constitutive Model Generation \\ 

\cite{zhang_machine_2021} & Inverse Material Design of Ionic Liquids for CO2 Capture \\ 
\midrule
\cite{tang_design_2021} & \multirow{2}{*}{Design of Metal-Organic Frameworks} \\ 
\cite{xiangyu_machine_2020}&  \\ 

\botrule
\end{tabular}
\end{center}
\end{table}

\paragraph{Electronics Design} In the design of electronic circuits, neural \gls{MCTS} is applied to solve routing problems in multiple cases \cite{chen_reinforcement_2022, he_circuit_2022, raina_learning_2022} (see \cref{tab:electronics_applications}), where it can outperform e.g. traditional A* based approaches \cite{he_circuit_2022}. 
Thacker et al. \cite{thacker_alphara_2021} further explore performing redundancy analysis in memory chips using neural \gls{MCTS}.

\begin{table}[h]
\begin{center}
\caption{Applications in electronic design.}\label{tab:electronics_applications}%
\begin{tabular}{@{}p{0.1\columnwidth}p{0.75\columnwidth}@{}}

\toprule
Source & Application \\ 
\midrule
\cite{chen_reinforcement_2022} & \multirow{3}{*}{Electronic Circuit Routing}  \\ 
\cite{he_circuit_2022} & \\ 
\cite{raina_learning_2022} &  \\ 
\midrule
\cite{thacker_alphara_2021} & Memory Chip Redundancy Analysis \\

\botrule
\end{tabular}
\end{center}
\end{table}

\paragraph{Energy Systems} Neural \gls{MCTS} finds many application in the operation of energy systems, especially in innovative grid concepts which aim to enable a more sustainable energy supply (see \cref{tab:energy_applications}). This includes optimizing the operation of residential microgrids in an online fashion \cite{shuai_online_2021} as well as non-intrusive load monitoring and identification \cite{jiang_reinforced_2022}. 

Some attention is also directed towards managing a grid consisting of renewable energy sources and battery systems, which absorb the former's fluctuations in power output. For instance, Al-Saffar et al. \cite{al-saffar_reinforcement_2020} devise a system to coordinate voltage regulation in a distributed energy network with battery systems at multiple locations, while \cite{wu_intelligent_2021} use neural \gls{MCTS} to address predictive maintenance problems in such systems. 

\begin{table}[h]
\begin{center}
\caption{Applications in energy systems.}\label{tab:energy_applications}%
\begin{tabular}{@{}p{0.1\columnwidth}p{0.75\columnwidth}@{}}

\toprule
Source & Application \\ 
\midrule
\cite{al-saffar_reinforcement_2020} & Voltage Regulation in Distributed Energy Systems \\ 
\cite{jiang_reinforced_2022} & Load Monitoring and Identification in Smart Grids \\ 
\cite{shuai_online_2021} & Residential Microgrid Scheduling \\ 
\cite{sun_hybrid_2021} & Electrical Transmission Network Self-Healing \\ 
\cite{tomin_concept_2020} & Artificial Dispatcher Intelligent Control System in Electric Networks\\ 
\cite{wu_intelligent_2021} & Predictive Maintenance Management for Battery Energy Storage Systems \\ 

\botrule
\end{tabular}
\end{center}
\end{table}

\paragraph{Production} Applications in production systems mainly concern themselves with various kinds of scheduling approaches. Here, the processing sequence of jobs or operations on different machines is to be determined to e.g. minimize the total time until all jobs have finished processing (see \cref{tab:production_applications}). Traditional \gls{RL} approaches are also increasingly being investigated for these types of problems \cite{zhang_learning_2020, gannouni_neural_2020, samsonov_manufacturing_2021, kemmerling_towards_2021}. A closer examination of the advantages and disadvantages of traditional \gls{RL} and neural \gls{MCTS} methods for scheduling approaches may be an interesting line of future research. 

Further applications in production include line buffer planning in car manufacturing \cite{gros_real-time_2020} as well as assembly planning in collaborative human-robot scenarios \cite{yu_mastering_2020}. 

\begin{table}[h]
\begin{center}
\caption{Applications in production systems.}\label{tab:production_applications}%
\begin{tabular}{@{}p{0.1\columnwidth}p{0.75\columnwidth}@{}}

\toprule
Source & Application \\ 
\midrule
\cite{gros_real-time_2020} & Production Line Buffer Planning \\ 
\cite{goppert_predicting_2021} & Dynamically Interconnected Assembly Systems Scheduling \\ 
\cite{kumar_production_2021} & Industrial Mining Production Scheduling \\ 
\cite{rinciog_sheet-metal_2020} & Sheet Metal Production Scheduling \\ 
\cite{wang_parallel_2020} & Parallel Machine Workshop Scheduling \\ 
\cite{yu_mastering_2020} & Collaborative Assembly \\

\botrule
\end{tabular}
\end{center}
\end{table} 

\paragraph{Combinatorial Optimization} While the scheduling problems described above are problems from the field of combinatorial optimization, the authors approach them from a production perspective and pay close attention to the details of their individual use cases. A second group of combinatorial optimization applications can be found in \cref{tab:optimization_applications}. Here, the problems are more abstract and investigated from a computer science lens. 

Combinatorial optimization problems share many similarities with combinatorial board games. In a reinforcement learning context, they are typically solved constructively by building a solution iteratively from scratch, or they are solved by improvement, i.e. by iteratively improving some existing solution. In both cases, the problem features inherently discrete time steps and an inherently discrete action space. Differences can be observed in that there is no obvious notion of winning or losing, but rather a sense of relative performance. In addition, combinatorial games feature a fixed board and a fixed set of game pieces. In, e.g. a traveling salesman problem, the equivalent of a board may be considered a graph with weighted edges connecting different cities. Such a graph, however, will vary, with each problem instance consisting of different cities to be traveled through. In machine scheduling problems, operations may be considered the game pieces. Depending on the exact problem formulation, each operation needs to be processed on a specific machine for a specific duration. An operation could therefore be described as a game piece which can be freely parameterized by properties such as the duration, contrary to pieces in typical games.

\begin{table}[h]
\begin{center}
\caption{Applications in combinatorial optimization problems.}\label{tab:optimization_applications}%
\begin{tabular}{@{}p{0.1\columnwidth}p{0.75\columnwidth}@{}}

\toprule
Source & Application \\ 
\midrule
\cite{huang_coloring_2019} & Graph Coloring \\ 
\cite{kim_solving_2022} & PBQP-based register allocation \\ 
\cite{oren_solo_2021} & Machine Scheduling, Vehicle Routing \\ 
\cite{wu_research_2021} & Bin Packing \\ 
\midrule
\cite{xing_graph_2020} & \multirow{2}{*}{Traveling Salesman Problem} \\ 
\cite{xing_solve_2020} &  \\ 
\midrule
\cite{xu_learning_2019}& \multirow{2}{*}{Highest Safe Rung Problem} \\ 
\cite{xu_learning_2020} & \\
\midrule
\cite{xu_towards_2022} & Quantified Boolean Formula Satisfaction \\

\botrule
\end{tabular}
\end{center}
\end{table}

\paragraph{Cloud \& Edge Computing} As before in the production domain, a primary application in cloud and edge computing concerns scheduling problems (see \cref{tab:cloud_applications}). Again, the scheduling problems are combinatorial optimization problems, but the authors' interests arise from the domain of cloud computing itself and the presented problems are less abstract. 

\begin{table}[h]
\begin{center}
\caption{Applications in cloud and edge computing.}\label{tab:cloud_applications}%
\begin{tabular}{@{}p{0.1\columnwidth}p{0.75\columnwidth}@{}}

\toprule
Source & Application \\ 
\midrule
\cite{chen_intelligent_2020} & Task Offloading for UAV Edge Computing \\ 
\cite{chen_iraf_2019} & Resource Allocation in Mobile Edge Computing \\ 
\midrule
\cite{cheng_smart_2019} & \multirow{2}{*}{Directed Acyclic Graph Task Scheduling} \\ 
\cite{hu_spear_2019} &  \\ 
\midrule
\cite{peng_lore_2022} & Cloud Workflow Scheduling \\ 

\botrule
\end{tabular}
\end{center}
\end{table}

\paragraph{Graph Navigation} 
Navigating a graph from a given node to a target node is a relevant task in many settings, but is gaining attention particularly in knowledge graph research. Here, a common task is knowledge graph completion, which involves the prediction of missing relations between individual entities \cite{wang_research_2022}. Graph navigation is an important sub-task of knowledge graph completion \cite{shen_m-walk_2018}, for which neural \gls{MCTS} has been investigated (see \cref{tab:graphapplications}) and been shown to outperform existing baselines \cite{shen_m-walk_2018}. 

\begin{table}[h]
\begin{center}
\caption{Applications on graphs.}\label{tab:graphapplications}%
\begin{tabular}{@{}p{0.1\columnwidth}p{0.75\columnwidth}@{}}

\toprule
Source & Application \\ 
\midrule
\cite{he_neurally-guided_2018} & Knowledge Graph Navigation \\ 
\cite{shen_m-walk_2018} & (Knowledge) Graph Navigation \\ 
\cite{wang_learning_2021} & Graph Navigation \\ 
\cite{wang_research_2022} & Knowledge Graph Completion \\ 

\botrule
\end{tabular}
\end{center}
\end{table}

\paragraph{Networking \& Communications}  Applications in networking and communications range from network function virtualization \cite{fu_policy_2021, kuai_fair_2022}, to network topology optimization \cite{meng_deep_2019, wang_research_2021, yang_research_2018, zou_research_2019, zou_deploying_2020, yin_optimal_2021}, to spectrum sharing in mobile networks with multiple radio access technologies \cite{challita_deep_2021, yan_smart_2018}. 

One notable example here is the use of neural \gls{MCTS} for intrusion defense in software defined networking scenarios \cite{gabirondo-lopez_towards_2021}. Here, the defense problem is actually modelled as a two-player game, which is an exception among mostly single player scenarios within the surveyed literature.

\begin{table}[h]
\begin{center}
\caption{Applications in networking and communications.}\label{tab:networking_applications}%
\begin{tabular}{@{}p{0.1\columnwidth}p{0.75\columnwidth}@{}}

\toprule
Source & Application \\ 
\midrule
\cite{fu_policy_2021} & Service Function Chain Deployment Problem in Network Function Virtualization \\ 
\cite{gabirondo-lopez_towards_2021} & Intrusion Defense in Software Defined Networking \\ 
\cite{kuai_fair_2022} & Network Function Virtualization Mapping and Scheduling \\ 
\cite{wang_solving_2022} & Virtual Network Embedding \\ 
\cite{wang_self-play_2022} & Ressource Assignment in OpenRAN Networks \\ 
\midrule
\cite{meng_deep_2019} & \multirow{6}{*}{\shortstack{Layout Optimization of Mobile Ad Hoc \\ Networks}} \\
\cite{wang_research_2021} &  \\ 
 \cite{yang_research_2018}&  \\ 
\cite{zou_research_2019} &  \\ 
\cite{zou_deploying_2020} &  \\ 
\cite{yin_optimal_2021} &  \\  
\midrule
\cite{challita_deep_2021} & Dynamic Spectrum Sharing of LTE and NR \\ 
\cite{sandberg_learning_2022} & Radio Resource Scheduling \\ 
\cite{shafieirad_meeting_2022} & Wireless Communication Scheduling \\ 
\cite{yan_smart_2018} & Multi-RAT Access \\ 
\cite{mo_deepmcts_2022} & MIMO Detection \\ 
\cite{chen_ipas_2021} & Pilot-power Allocation for MIMO Systems \\ 

\botrule
\end{tabular}
\end{center}
\end{table}   

\paragraph{Autonomous Driving \& Motion Planning} 
Autonomous driving applications make up a comparatively large group (\cref{tab:driving_applications}), including general motion planning tasks \cite{ha_vehicle_2020, lei_kb-tree_2021, paxton_combining_2017, weingertner_monte_2020}, motion planning tasks in autonomous parking scenarios \cite{song_data_2020, zhang_reinforcement_2020}, and motion planning tasks in multi-agent settings \cite{riviere_neural_2021, skrynnik_hybrid_2021}. More specialised tasks such as lane keeping \cite{kovari_design_2020}, overtaking \cite{mo_safe_2022}, and higher level decision making during autonomous driving \cite{hoel_combining_2020} are considerd as well. 

Such problems are often fundamentally different from combinatorial games. For instance, Weingertner et al. \cite{weingertner_monte_2020} consider a motion planning problem, in which the acceleration of a vehicle is controlled along a predetermined path. In its natural formulation, such a problem requires selecting continuous actions in continuous time. To apply neural \gls{MCTS}, both time and action space are discretized. The resulting solution demonstrates good performance and outperforms A* search, pure deep learning approaches, and model predictive control.

\begin{table}[h]
\begin{center}
\caption{Applications in autonomous driving, as well as path and motion planning.}\label{tab:driving_applications}%
\begin{tabular}{@{}p{0.1\columnwidth}p{0.75\columnwidth}@{}}

\toprule
Source & Application \\ 
\midrule
\cite{song_data_2020} & \multirow{2}{*}{Motion Planning in Autonomous Parking} \\ 
\cite{zhang_reinforcement_2020} &  \\ 
\midrule
\cite{ha_vehicle_2020} & \multirow{4}{*}{Motion Planning in Autonomous Driving} \\ 
 \cite{lei_kb-tree_2021} &  \\ 
\cite{paxton_combining_2017} &  \\ 
\cite{weingertner_monte_2020} & \\ 

\midrule
\cite{chen_driving_2020} & Driving Maneuver Prediction \\ 
\cite{hoel_combining_2020} & Tactical Decision Making \\ 
\cite{kovari_design_2020} & Lane Keeping Tasks \\ 
\cite{mo_safe_2022} & Overtaking Tasks \\ 
\cite{riviere_neural_2021} & Multi-Robot Motion and Path Planning \\ 
\cite{skrynnik_hybrid_2021} & Multi-Vehicle Motion and Path Planning \\ 

\botrule
\end{tabular}
\end{center}
\end{table}    

\paragraph{Natural Language Processing} In \cref{tab:nlpapplications}, several applications of conversational agents are shown, in which agents assist users in completing tasks \cite{wang_task-completion_2020}, negotiate with users to divide a given set of resources \cite{jang_bayes-adaptive_2020}, and try to convince users of a certain view by framing messages in different ways \cite{carfora_dialogue_2020}. While humans are difficult to simulate as conversational partners explicitly, models that approximate narrow conversational behaviour of humans can be trained on historical data and then utilized as part of the tree search \cite{carfora_dialogue_2020, wang_task-completion_2020}. 

Natural language processing is itself a diverse field, in which topics such as sentiment analysis \cite{dai_reinforcement_2021} and named entity recognition \cite{lao_name_2019} are being addressed with neural \gls{MCTS}. 
\begin{table}[h]
\begin{center}
\caption{Applications in Natural Language Processing.}\label{tab:nlpapplications}%
\begin{tabular}{@{}p{0.1\columnwidth}p{0.75\columnwidth}@{}}

\toprule
Source & Application \\ 
\midrule
\cite{carfora_dialogue_2020} & Personalized Messaging \\ 
\cite{jang_bayes-adaptive_2020} & Negotiation Dialogues \\ 
\cite{wang_task-completion_2020} & Task-Completion Dialogues \\ 
\midrule
\cite{dai_reinforcement_2021} & Sentiment Analysis \\ 
\cite{he_text_2018} & Text Matching \\ 
\cite{lao_name_2019} & Named Entity Recognition \\

\botrule
\end{tabular}
\end{center}
\end{table}    
   
\paragraph{Machine Learning} \gls{MCTS} guided by machine learning models can in turn be used in certain machine learning tasks (see \cref{tab:mlapplications}). For instance \cite{huang_distilling_2018} and \cite{wang_channel_2022} apply neural \gls{MCTS} to reduce the size of neural networks by network distillation in the former and \gls{CNN} filter pruning in the latter case. 

Lu et al. \cite{lu_incorporating_2021} further approach the task of symbolic regression with neural \gls{MCTS}. Here, instead of solving regression tasks by adjusting the coefficients of a e.g. a linear or polynomial function, the terms of a function themselves (e.g. sinusoids, square operations, constants) are determined and connected through mathematical operators such as addition and divison.  In \gls{MCTS}, the full expression of a function can be built up step by step.
 
\begin{table}[h]
\begin{center}
\caption{Applications in machine learning.}\label{tab:mlapplications}%
\begin{tabular}{@{}p{0.1\columnwidth}p{0.75\columnwidth}@{}}

\toprule
Source & Application \\ 
\midrule
\cite{huang_distilling_2018} & Neural Network Distillation \\ 
\cite{wang_channel_2022} & CNN Filter Pruning \\ 
\midrule
\cite{chen_costly_2021} & Costly Feature Classification \\ 
\cite{lu_incorporating_2021} & Symbolic Regression \\ 
\cite{zou_reinforcement_2019} & Recommender Systems \\ 

\botrule
\end{tabular}
\end{center}
\end{table}

\paragraph{Computer Science} Computer Science offers a wide range of opportunities for the application of neural \gls{MCTS} (see \cref{tab:csapplications}), many of which are presented in separate sections. Others do not warrant their own section due the small amount of publications in their specific niche, but are nevertheless interesting. The amount of publications this applies to demonstrates the wide applicability of neural \gls{MCTS}. 

One notable example is AlphaTensor \cite{fawzi_discovering_2022}, where neural \gls{MCTS} is used to find efficient algorithms for matrix multiplication. Others include the optimization of database queries \cite{zhang_alphajoin_2020}, the recovery of sparse signals \cite{chen_deeppursuit_2021, zhong_learning_2019}, and various applications in quantum computing \cite{chen_optimizing_2022, dalgaard_global_2020, sinha_qubit_2022}. 

\begin{table}[h]
\begin{center}
\caption{Applications in computer science.}\label{tab:csapplications}%
\begin{tabular}{@{}p{0.1\columnwidth}p{0.75\columnwidth}@{}}

\toprule
Source & Application \\ 
\midrule
\cite{fawzi_discovering_2022} & Matrix Multiplication Algorithm Discovery \\ 
\cite{shao_alphaseq_2020} & Sequence Discovery \\ 
\cite{xu_--fly_2022} & Model Checking \\ 
\cite{zhang_alphajoin_2020} & Database Query Optimization \\ 
\cite{zhang_construction_2018} & Low-Density Parity-Check Code Construction \\ 
\midrule
\cite{chen_deeppursuit_2021} & \multirow{2}{*}{Sparse Signal Recovery} \\ 
\cite{zhong_learning_2019} &  \\ 
\midrule
\cite{gauthier_deep_2020} & \multirow{3}{*}{Automatic Theorem Proving} \\ 
\cite{zombori_prolog_2020} & \\ 
\cite{zombori_role_2021} &  \\ 
\midrule
\cite{chen_optimizing_2022} & Quantum Annealing Schedule Optimization \\ 
\cite{dalgaard_global_2020} & Quantum Dynamics Optimization \\ 
\cite{sinha_qubit_2022} & Qubit Routing \\ 
\botrule
\end{tabular}
\end{center}
\end{table}

\begin{table}[h]
\begin{center}
\caption{Applications in various other fields.}\label{tab:variousapplications}%
\begin{tabular}{@{}p{0.1\columnwidth}p{0.75\columnwidth}@{}}

\toprule
Source & Application \\ 
\midrule
\cite{yang_reinforcement_2020} & Active Space Multi-Debris Removal \\ 
\cite{zhang_reinforcement-learning-based_2022} & Configuration of Cellular Satellites \\ 
\midrule
\cite{gaafar_reinforcement_2019} & \multirow{2}{*}{Cognitive Radar Task Scheduling} \\ 
\cite{shaghaghi_resource_2019} &  \\ 
\midrule
\cite{bai_hierarchical_2022} & Object Rearrangement \\ 
\cite{wang_scene_2020} & Scene Arrangement Planning \\ 
\midrule
\cite{silva_dynamic_2021} & \multirow{2}{*}{User Interface Optimization} \\ 
\cite{todi_adapting_2021} &  \\ 
\midrule
\cite{sadeghnejad-barkousaraie_reinforcement_2021} & \multirow{2}{*}{Radiotherapy Beam Orientation Selection} \\ 
\cite{sadeghnejad_barkousaraie_using_2019}  & \\ 
\midrule
\cite{audrey_deep_2019} & Fluid Structure Topology Optimization \\ 
\cite{raina_learning_2022} & Truss Design \\ 
\midrule
\cite{dai_aoi-minimal_2022} & Mobile Crowdsensing \\ 
\cite{feng_electromagnetic_2021} & Electromagnetic Situation Analysis \\ 
\cite{fong_model-based_2021} & Robotic Manipulation \\ 
\cite{ganapathi_subramanian_combining_2018} & Wildfire Spread Prediction \\ 
\cite{kovari_policy_2020} & Pneumatic Actuator Control \\ 
\cite{li_rich-text_2021} & Document Style Reverse Engineering \\ 
\cite{wang_timetable_2019} & Railway Timetable Rescheduling \\ 

\botrule
\end{tabular}
\end{center}
\end{table}    

Finally, \cref{tab:variousapplications} shows applications in various other fields that do not fit into any of the previous categories. These feature a diverse set of problems including the optimization of user interfaces \cite{silva_dynamic_2021, todi_adapting_2021}, control of a pneumatic actuator \cite{kovari_policy_2020}, as well as design tasks for trusses \cite{raina_learning_2022} and fluid structures \cite{audrey_deep_2019}. As in many examples here, similar design tasks are also being approached with traditional \gls{RL} \cite{fricke_investigation_2023}. In future studies, direct comparisons of traditional \gls{RL} and neural \gls{MCTS} methods on specific problems may help to decide what approach is preferable under which conditions. \\
\newline
 In summary, the applications described in the above sections originate in a variety of different disciplines including chemistry, medicine, computer science, mathematics, and electrical engineering. The types of problems include optimization tasks of various kinds, control problems, generative design tasks, and many others. Clearly, neural \gls{MCTS} shows wide applicability beyond combinatorial games, to problems which in part share and do not share the properties of games.  

\subsection{Application Characteristics}
Like games, the problems surveyed here can be formulated as a \gls{MDP}. Playing combinatorial games involves choosing discrete actions at discrete time-steps in a finite horizon setting, i.e. games are episodic with well-defined terminal conditions. Rewards are typically sparse and correspond to a small set of possible game outcomes: loss (-1), draw (0), and win (1). While the state transitions of each individual player are typically deterministic, the presence of a second player introduces uncertainty about the states which will be encountered at the next turn. While many of the applications surveyed here share many of these properties, neural \gls{MCTS} is also applied to applications which differ from combinatorial games in one or multiple dimensions. 
\paragraph{Time}
Many settings do not have a turn-based nature, but allow for the execution of actions at arbitrary points in time, i.e. time often has a continuous nature. This does not appear to hinder the application of neural \gls{MCTS}, as many authors simply discretize the time dimension in their problem formulation \cite{bai_hierarchical_2022, fong_model-based_2021, ganapathi_subramanian_combining_2018, ha_vehicle_2020, huang_distilling_2018, kovari_policy_2020, kovari_design_2020, lao_name_2019, paxton_combining_2017, song_data_2020, weingertner_monte_2020, zhang_reinforcement-learning-based_2022}.

\paragraph{Finite \& Infinite Horizons}
Like combinatorial games, most of the applications surveyed here consist of a finite horizon problem, i.e. the problem is solved in episodes of finite length. In some cases, the natural formulation of the problem features an infinite horizon. To apply neural \gls{MCTS}, episodes can then be created artificially by setting a maximum number of steps after which the episode always terminates, as is done in \cite{kumar_production_2021, shuai_online_2021, todi_adapting_2021}.

\paragraph{Transitions}
Many of the surveyed problems are of a completely deterministic nature, which is a fundamental difference compared to the combinatorial games domain. In such cases, the tree search may be modified to take advantage of the deterministic transitions (see \cref{subsec:guided_selection} for more details).  

Nevertheless, some problem formulations with stochastic state transitions can be observed \cite{gabirondo-lopez_towards_2021, goppert_predicting_2021, kumar_production_2021, meng_deep_2019, shafieirad_meeting_2022, todi_adapting_2021, wang_task-completion_2020, yan_smart_2018}. 

\paragraph{Rewards}
The reward structure of typical problem settings often does not share the simplicity of the reward function present in games. Instead of a set with  two or three distinct reward values, rewards are typically given on a continuum corresponding to the quality of the obtained solutions. Often, the rewards are not even clearly bounded on one or both sides (see e.g. \cite{bai_hierarchical_2022, hu_spear_2019, mao_learning_2021, segler_planning_2018, shuai_online_2021, wang_task-completion_2020, yu_mastering_2020, zhang_reinforcement-learning-based_2022}). 

In some cases, the reward is transformed by self-play inspired mechanisms. This will be investigated in more detail in \cref{subsec:selfplay}. 

While the majority of surveyed problems feature some kind of sparse reward at the end of an episode, in some cases, more fine-grained rewards after each action are incorporated into the tree search \cite{fawzi_discovering_2022, hoel_combining_2020, sridharan_deep_2022}. 

\paragraph{Action Spaces}
\gls{MCTS} naturally lends itself well to discrete action spaces, as is the case in combinatorial board games. While modifications of (neural) \gls{MCTS} for continuous action spaces exist \cite{moerland_a0c_2018}, the vast majority of applications surveyed here exhibit discrete action spaces. Notable exceptions are the approaches of Lei et al.  \cite{lei_kb-tree_2021} and Paxton et al. \cite{paxton_combining_2017}

Further, Raina et al. \cite{raina_learning_2022} apply a hierarchical reinforcement learning approach, in which neural \gls{MCTS} is used for an overarching set of discrete actions while subsequent, continuous actions are determined by another mechanism. 

Finally, it is always possible to discretize a naturally continuous action space. While this reduces the amount of precision with which actions can be chosen, some applications can nevertheless be successfully approached in this manner \cite{fawzi_discovering_2022, shuai_online_2021}. 

\paragraph{State Spaces} 
While the exact characteristics of state spaces depend not only on the underlying problem, but also on how the problem is modelled, games such as Go have a well-defined, regular board, which is helpful in formulating a state space. In Go, the board of fixed size consisting of cells which are positioned in spatial relation to each other lends itself well to processing by a \gls{CNN}. The problems surveyed here feature a diverse range of state spaces which are processed by different kinds of neural networks. Next to \gls{CNN}s, the employed neural networks include recursive neural networks such as long short-term memory networks \cite{bai_hierarchical_2022,erikawa_mermaid_2021,fong_model-based_2021,gauthier_deep_2020,he_neurally-guided_2018,he_text_2018,huang_coloring_2019,shuai_online_2021,todi_adapting_2021,wang_scene_2020} and gated recurrent units \cite{dai_reinforcement_2021,ganapathi_subramanian_combining_2018,jang_bayes-adaptive_2020,lao_name_2019,ma_structure-based_2021,srinivasan_artificial_2021,tang_design_2021,wang_meta-modeling_2019,wang_learning_2021,xu_towards_2022,yang_chemts_2017,zou_reinforcement_2019}, as well as graph neural networks \cite{ishida_ai-driven_2022,kim_solving_2022,lei_kb-tree_2021,li_structure-based_2021,oren_solo_2021,peng_lore_2022,sinha_qubit_2022,sridharan_deep_2022,sv_multi-objective_2022,wang_learning_2021,xing_graph_2020,xing_solve_2020,xu_towards_2022,zombori_role_2021}. 
Less frequent types include transformers \cite{fawzi_discovering_2022,qian_alphadrug_2022,sandberg_learning_2022} and the DeepSet architecture \cite{yin_optimal_2021}. In many cases, a simple multi-layer perceptron is sufficient \cite{challita_deep_2021,chen_intelligent_2020,chen_iraf_2019,chen_ipas_2021,chen_optimizing_2022,dalgaard_global_2020,gabirondo-lopez_towards_2021,genheden_aizynthfinder_2020,gros_real-time_2020,goppert_predicting_2021,he_circuit_2022,hu_spear_2019,kovari_policy_2020,kovari_design_2020,kumar_production_2021,paxton_combining_2017,rinciog_sheet-metal_2020,segler_planning_2018,segler_towards_2017,shuai_post-storm_2023,thakkar_datasets_2020,wang_task-completion_2020,wang_towards_2020,weingertner_monte_2020,xu_--fly_2022,zhang_reinforcement_2020,zhang_alphajoin_2020,zhong_learning_2019}.

The chosen architecture and its depth will to some degree determine what kind of hardware is required to train a neural \gls{MCTS} approach. 

\subsection{Hardware Requirements}
The training of AlphaZero involved more than 5000 tensor processing units \cite{silver_general_2018}. One might hence question whether the application of neural \gls{MCTS} is a viable option for researchers and practitioners who do not have access to resources of that magnitude. 

Some of the publications we review do report the usage of significant resources. For example, Huang et al. \cite{huang_coloring_2019} use up to 300 NVIDIA 1080Ti and 2080Ti GPUs during training. The majority of the reported hardware, however, is not out of reach for typical organizations and even private individuals. Genheden et al. \cite{genheden_aizynthfinder_2020} report that they use a single Intel Xeon CPU with a single NVIDIA 2080Ti GPU on a machine with 64 GB memory. Many others use a single high-end consumer CPU and GPU  \cite{kuai_fair_2022, kumar_production_2021, sadeghnejad-barkousaraie_reinforcement_2021, shao_alphaseq_2020, zhong_learning_2019}.

On the lower end, some researchers even use consumer notebooks to train neural \gls{MCTS} methods \cite{hu_spear_2019, xu_towards_2022, yang_reinforcement_2020}. 

Clearly, hardware requirements vary by application and the complexity of the employed neural networks. While it is difficult to predict what level of hardware is required for a given application and desired solution quality, it is clear that moderately powerful hardware can be successfully utilized in many applications.

\section{Neural MCTS Methodologies}
\label{sec:results_nmcts_methodologies}

After gaining an overview of the breadth of possible neural \gls{MCTS} applications in the previous section, we now turn our attention to the design of neural \gls{MCTS} approaches as they were encountered during the review. 

\subsection{Guidance Network Training}
Before delving into the inner mechanisms of neural \gls{MCTS}, we content ourselves with the knowledge that learned policy and value networks are used to guide the tree search in some way. In the following, we first dedicate some attention to the training procedure used in AlphaZero \cite{silver_general_2018}, before discussing alternatives found during the review.  

\paragraph{Policy improvement by MCTS}
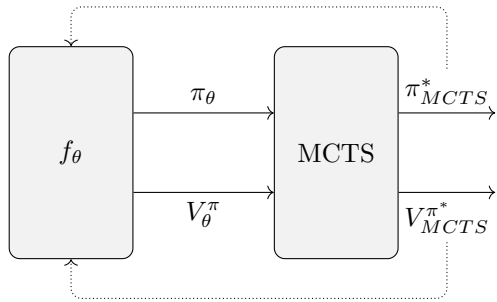
\begin{figure}
    \centering
    \input{Fig3}
    \caption{\gls{MCTS} as a policy improvement operator. The learned policy and value function are used to guide the tree search, which then produces an improved policy $\pi_{MCTS}^{*}$ and value estimate $V_{MCTS}^{\pi^*}$ for a given state. As visualized by the dotted lines, $\pi_{MCTS}^{*}$ and $V_{MCTS}^{\pi^*}$ can then also be used as training targets for the neural network.}
    \label{fig:mcts_policy_improvement}
\end{figure}
In AlphaZero \cite{silver_general_2018}, a learned policy is iteratively improved by guiding an \gls{MCTS} search and in turn using the search results to improve the learned policy (see \cref{fig:mcts_policy_improvement}). We refer to this procedure as \emph{policy improvement by \gls{MCTS}}. More concretely, the learned policy $\pi_\theta$ guides the tree search in one or multiple of the search phases and a new policy $\pi_{MCTS}$ for the state under consideration is obtained after a given number of \gls{MCTS} simulations. Since this new policy is typically stronger than the initial, learned one, it can be used as a training target for the policy network. More precisely, the policy network and \gls{MCTS} produce policy probability vectors $\mathbf{p}_\theta$ and $\mathbf{p}_{MCTS}$ for a given state, where the former can be seen as the actual prediction and the latter as the prediction target. These can then be used in a cross-entropy loss function to train the policy network: $L_{CE} = - \ \mathbf{p}_{MCTS}^T \ \log \  \mathbf{p}_\theta$. Accordingly, if a value function is learned alongside the policy, its value estimates are adjusted in the direction of those found by \gls{MCTS} by using the \gls{MSE} as a loss function: $L_{MSE} = (v_{MCTS}-v)^2$. Both terms are typically combined with a regularization term into a single loss function 

\begin{equation}
\label{eq:pmctsloss}
  L = (v_{MCTS}-v)^2 - \ \mathbf{p}_{MCTS}^T \ \log \  \mathbf{p}_\theta + c \ \| \theta \|^2
\end{equation}

The vast majority of the articles we reviewed that improve a learned policy by MCTS use the loss function given in \cref{eq:pmctsloss}. We find two exceptions to this, one in which the Kullback–Leibler divergence is used instead of the cross-entropy \cite{chen_costly_2021} and one in which the Kullback–Leibler divergence is also used instead of the cross-entropy, but a quantile regression distributional loss is additionally used instead of the \gls{MSE} \cite{fawzi_discovering_2022}. It may be worth investigating what effect these loss functions have on training, but the combination of cross-entropy and \gls{MSE} clearly emerges as the default choice during our review. 

\begin{figure*}
    \centering
    \includegraphics[width=0.75\textwidth]{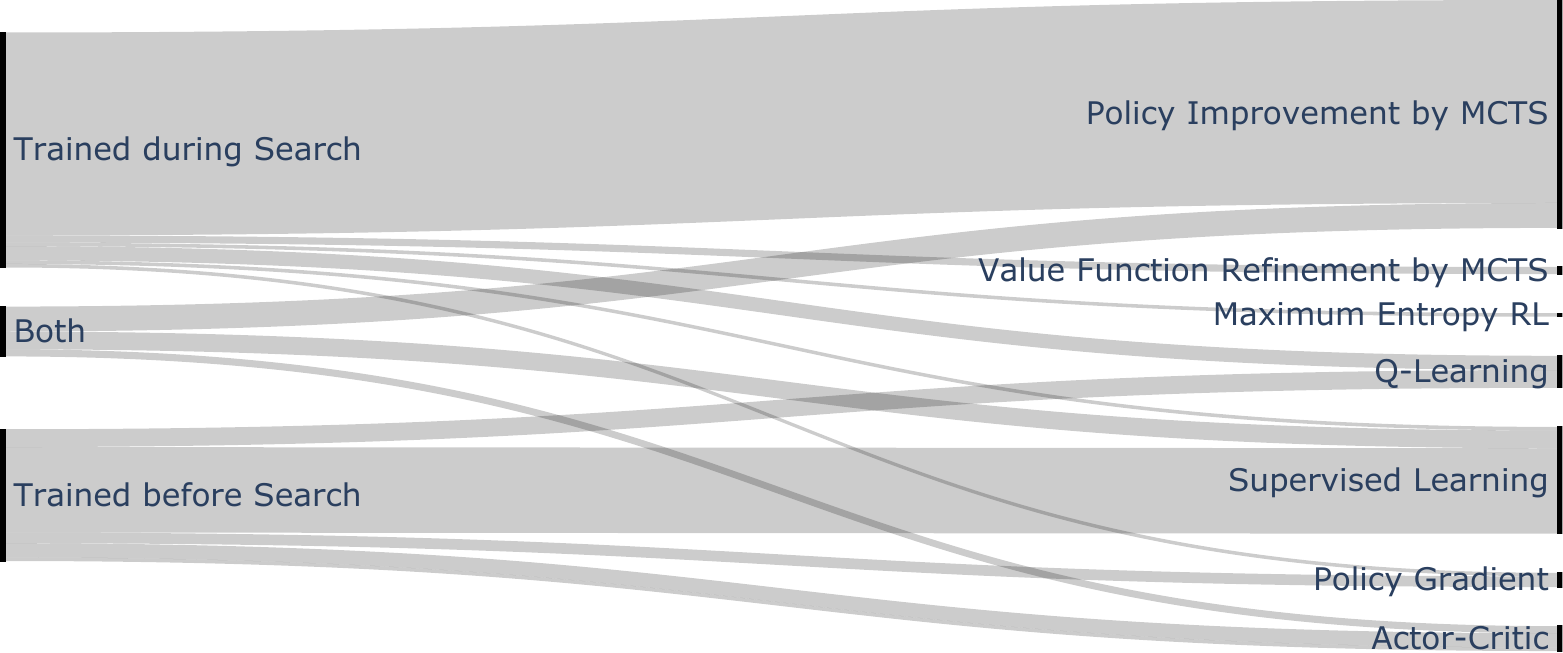}
    \caption{Depending on the approach, policy and value networks are trained before the search, during the search, or both. Depending on this choice, different training methods are employed. Policy gradient and Actor-Critic are families of algorithms that encompass multiple specific algorithms.}
    \label{fig:training_methods}
\end{figure*}

\paragraph{Alternative training approaches}
We find two broad groups of training approaches during our review: (1) training before the networks are used to guide the tree search and (2) training during the tree search, i.e. iteratively performing tree search and using its results for training. In group (1), training is facilitated either by supervised learning on labelled examples or by classical reinforcement learning algorithms in the policy-based, value-based, and actor-critic families. In group (2), the dominant approach is policy improvement by \gls{MCTS}, but some variations on this exist. Finally, both approaches can be combined by first performing what can be considered a pre-training and then training further during the tree search. This is sometimes referred to as a warm-start. 

\Cref{fig:training_methods} shows the distribution of these different approaches as found in the surveyed publications and which specific methods are employed in each approach. A large portion of authors train their networks during the search by using policy improvement by \gls{MCTS}. In some cases, training during the search occurs by other methods such as Q-Learning \cite{al-saffar_reinforcement_2020,gros_real-time_2020,shen_m-walk_2018,wang_learning_2021} and Maximum Entropy \gls{RL} \cite{zombori_role_2021} on the \gls{MCTS} trajectories. Instead of improving the policy (alongside the value function) by \gls{MCTS}, it is also feasible to refine the value function individually \cite{li_rich-text_2021,xing_solve_2020}, without any learned policy. We term this approach \emph{value function refinement by \gls{MCTS}} here. Likewise, while policy improvement by \gls{MCTS} typically trains both a policy and a value function, sometimes the policy is also trained in isolation \cite{shaghaghi_resource_2019}.

When trained before the tree search, networks are most often trained by supervised learning on labelled demonstrations. Q-Learning \cite{oren_solo_2021,paxton_combining_2017,wang_scene_2020,wang_task-completion_2020,weingertner_monte_2020}, policy gradient \cite{dieb_optimization_2020,hu_spear_2019,kovari_policy_2020,kuai_fair_2022,lu_incorporating_2021,peng_lore_2022}  and actor-critic methods \cite{chen_costly_2021,he_circuit_2022,wang_channel_2022,wang_parallel_2020} are also employed. 

When both training phases are performed, the most common approach is to combine supervised pre-training with policy improvement by \gls{MCTS}  \cite{chen_optimizing_2022,rinciog_sheet-metal_2020,sun_hybrid_2021,wang_towards_2020,zhang_reinforcement-learning-based_2022}.

Overall, there is some variety in the employed training approaches, but the dominant strategies are supervised training before the search and policy improvement by \gls{MCTS}, sometimes combined in one approach as in the original AlphaGo publication \cite{silver_mastering_2016}. In a combinatorial game setting, policy improvement by \gls{MCTS} requires some mechanism by which the opponent's moves are generated. In the following, such a mechanism and its relevance for applications beyond games are discussed. 

\subsection{Self-play Beyond Games}
\label{subsec:selfplay}
One of the components leading to the success of AlphaGo is the concept of self-play \cite{silver_mastering_2016}. In a self-play setting, opponents in a multi-player game are controlled by (some version of) the same policy, i.e. the policy plays against itself \cite{hernandez_generalized_2019}. Learning in such a scenario has the advantage that the policy always faces an opponent of comparable skill, which evolves as the training progresses. However, since many non-game-playing applications have an inherently single-player nature, the role of self-play beyond games is not obvious. 

To gain a clearer understanding of the applicability of self-play in such cases, we surveyed its usage among the publications included in our review. During this process, it became apparent that many authors use the term self-play, but that the meaning of the term varies. This may be due to the lack of an accepted, standardized definition. In the following, we first delineate different meanings of the term we encountered and then report the usages of different versions of self-play in our review. 

In a two-player setting, the term self-play can be understood intuitively. As used in the original AlphaGo publication \cite{silver_mastering_2016}, self-play entails that the policy currently being learned plays a game against some older version of itself in a two-player turn-based setting. This means that this other version of the current policy is being used to generate new states by playing every other turn of the game, as well as to obtain the final reward of the game. 

In single-player settings, the term self-play is also often used, but its meaning is less obvious. In a single-player setting, the \emph{state generating} property described above is not applicable, since the state transitions do not depend on the actions of another player. Generating a new state requires only the current state and the agent's action. The \emph{reward generating} property described above, however, is applicable if the reward function is designed accordingly. If the reward is not simply dependent on the performance of the current policy, but on the relative performance compared to some prior version of the policy, the reward generating property of self-play is transferred to the single-player setting. In other words, the process of a policy trying to beat its own high score has similarities to the concept of self-play in two-player settings. While this is sometimes also called self-play, Mandhane et al. \cite{mandhane_muzero_2022} introduce the term \emph{self-competition} for this type of approach. In the remainder of this review, we will adopt this term and reserve \emph{self-play} for multi-player settings to avoid confusion. While simple versions of self-competition can be implemented trivially, Laterre et al. \cite{laterre_ranked_2018} introduced a more substantiated form of self-competition named \emph{ranked reward}, followed by the approaches of Mandhane et al. \cite{mandhane_muzero_2022} and Schmidt et al. \cite{schmidt_self-play_2019}. 

However, many authors claim to implement self-play without obviously applying any of the concepts described above (see e.g. \cite{raina_learning_2022, thacker_alphara_2021}). While it is hard to be certain about what is meant in such instances, we suspect that two further concepts are sometimes termed self-play in the literature. The first is the practice of keeping track of the best policy by evaluating the current policy against the previous best one. If the current policy can outperform the previous best one on some defined set of problems, it replaces the currently saved best policy. This simply serves the purpose of having access to the best policy after training completes, since training does not necessarily improve the policy monotonically. In such cases, the outcomes of evaluation are not used as rewards to train the policy. Consequently, no learning follows from the policy playing against another version of itself, i.e. it is not a mechanism by which the current policy is improved, but merely evaluated. 

The last concept, which we suspect is sometimes described as self-play is policy improvement by \gls{MCTS} as introduced above (see e.g. \cite{chen_iraf_2019}). 

To be clear, we will use the term self-play only to describe multi-player cases where the state generating property as well as the reward generating property hold, and the term self-competition only for single-player cases where the reward generating property holds. 

While we would ideally like to report the usage of self-play and self-competition for all publications included in our review, we refrain from doing so when the terms are used ambiguously and instead only report a selection of notable examples where their meaning has been clearly established. 
\paragraph{Self-play}
Actual self-play appears to be fairly rare in the non-game-playing literature. We can only attribute the use of self-play to a single work \cite{feng_electromagnetic_2021}, in which a problem is modelled as a two-player game and a policy learned by self-play. In some other cases, problems are modelled as two-player games as well, but the resulting games are asymmetric, i.e. the players have different action spaces and hence require different policies  \cite{gabirondo-lopez_towards_2021,xu_learning_2019,xu_learning_2020,xu_--fly_2022,xu_towards_2022}. In such cases, two different neural networks each learn a policy.

\paragraph{Self-competition}
In terms of self-competition, we observe instances of ranked reward  \cite{sv_multi-objective_2022,wang_learning_2021,wang_self-play_2022} as well as naive approaches (see \cref{tab:self_competition}). In a naive approach, the performance of the current policy on the current problem is simply evaluated as some score and compared against the score of the best policy observed up to this point on the same problem. If the current policy's score is better, the game is won ($r=1$), if it is worse, the game is lost ($r=-1$), and if it is equivalent, the outcome is a draw ($r=0$) \cite{huang_coloring_2019,kim_solving_2022}. A variation of this is to not use the best policy, but to evaluate against the average score of a group of saved policies \cite{wang_meta-modeling_2019}. 

In one case, a naive approach, as described above, is applied, but instead of a past version of the policy, a second, completely independent policy is learned and the two policies continually compete against each other \cite{audrey_deep_2019}.

While self-competition can be used to generate rewards based on the relative performance of the policy, this is not strictly necessary, as the absolute performance can be used to compute rewards just as well. One benefit of self-competition may simply be having a reward in a clearly defined range of $[-1, 1]$ or similar, as optimal choices of \gls{MCTS} hyperparameters depend on this range \cite{browne_survey_2012}. However, there appear to be benefits beyond this, as the ranked reward approach has been shown to outperform agents trained using a standard reward in the range $[0, 1]$ \cite{laterre_ranked_2018}. Whether this is the case for the naive self-competition approaches as well is unclear. 

\begin{table}[h]

\begin{center}
\caption{Self-competition}\label{tab:self_competition}%
\begin{tabular}{@{}p{0.3\columnwidth}p{0.55\columnwidth}@{}}

\toprule
Type & Publications \\ 
\midrule
Ranked reward & \cite{sv_multi-objective_2022}, \cite{wang_learning_2021}, \cite{wang_self-play_2022} \\ 
Naive (best) & \cite{huang_coloring_2019}, \cite{kim_solving_2022} \\
Naive (average) & \cite{wang_meta-modeling_2019} \\

\botrule
\end{tabular}
\end{center}
\end{table}

\subsection{Guided Selection}
\label{subsec:guided_selection}

The previous sections argue that \gls{MCTS} functions as a policy improvement operator. We now explore the mechanisms of this policy improvement, i.e. the inner workings of neural \gls{MCTS}. A search iteration in \gls{MCTS} begins with the selection phase, in which actions are iteratively chosen starting from the root state until a leaf node is encountered. As described in \cref{subsec:mcts}, the choice of action is determined by a tree policy, which generally takes the form 
\begin{equation}
    a = \argmax_a{Q(s,a) + U(s,a)}
\label{eq:tree_policy}
\end{equation}

where $Q(s,a)$ encourages exploitation of known high value actions, while $U(s,a)$ encourages exploration of the search tree. Variations exist both in the exact formulation of \cref{eq:tree_policy} and in how individual terms of the equation are determined, i.e. by learned policies and value functions or by conventional means. We investigate each aspect individually in the following. 

\paragraph{Tree policy formulations}
The tree policies encountered during the review are usually based on some version or extension of the \gls{UCT} rule, but some variation in the exact formulation of the rule, especially in the exploration part, can be observed. 

\begin{figure}
    \centering
    \includegraphics[width=\columnwidth]{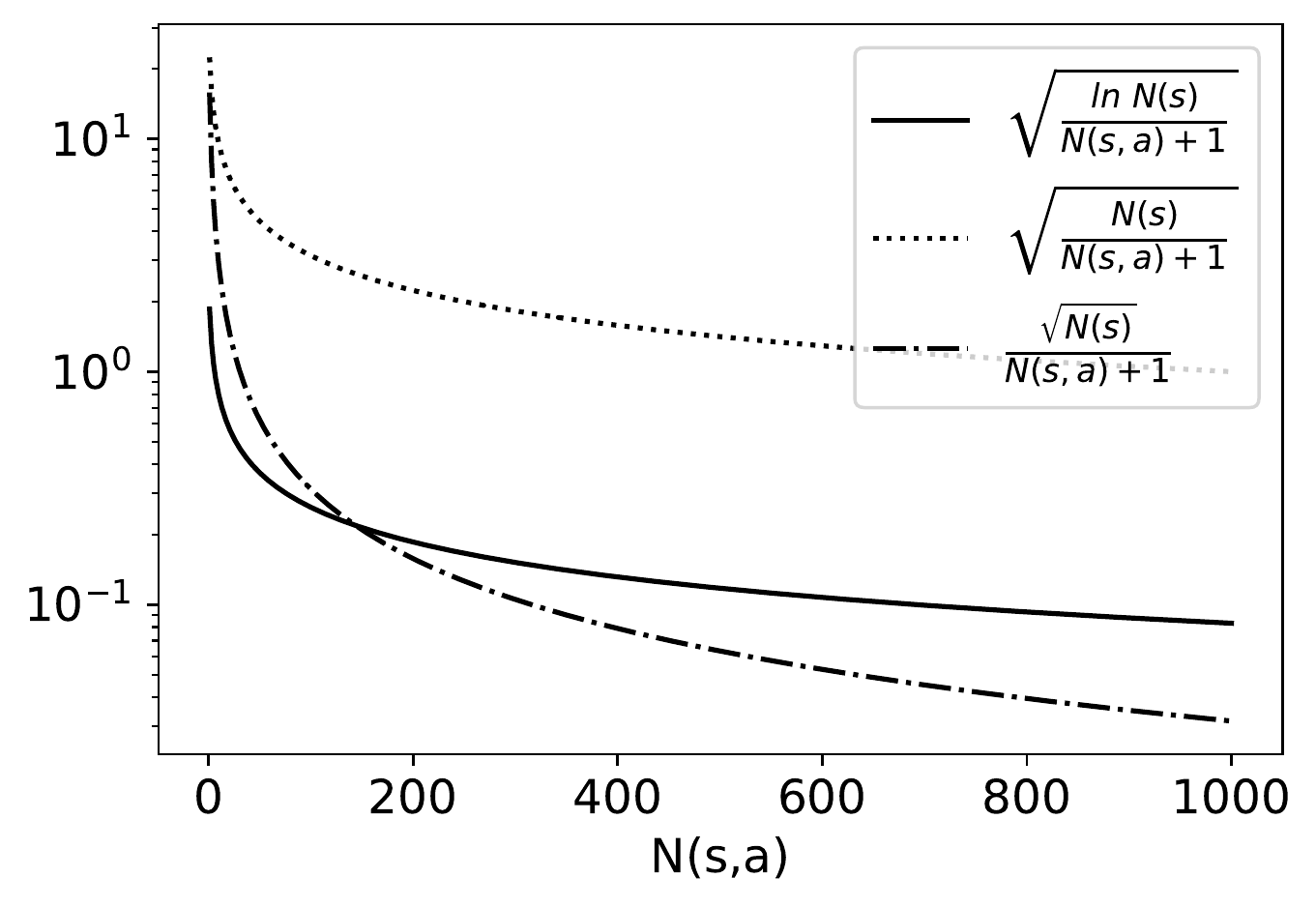}
    \caption{Different UCT-style fractions with a fixed $N(s) = 1000$. Note that the vertical axis is logarithmic and shows the value the expressions in the legend produce for different $N(s,a)$.}
    \label{fig:puct_uct}
\end{figure}

We provide an overview of variations of $U(s,a)$ identified during our review in \cref{tab:uct_variations}. While compiling the table, we modified the exact formulations reported in individual publications to arrive at a consistent notation. To this end, we assumed that all reported logarithms are natural logarithms and that $N(s) = \sum_b N(s,b)$, i.e. $N(s)$ refers to the visit count of all the children in state $s$, while $N(s, a)$ refers to the visit count of action $a$ in state $s$. The exploration constant, sometimes given as $c_{uct}$, $c_{puct}$ or similar, is simply referred to as $c$ in this review. $P(s, a)$ represents some prior probability of choosing action $a$ in state $s$, whether it be given by a learned policy or obtained by other means. 

\begin{table*}[h]

\begin{center}
\caption{Variations of tree policies based on \gls{UCT}. Different groups of selection formulae are divided by two horizontal lines and each member of a group has a variant number. The first member of a group is always the most frequently observed one, not necessarily the original formulation of the group.}\label{tab:uct_variations}%
\renewcommand{\arraystretch}{2}
\begin{tabular}{@{} c  c  c  p{7cm}@{}}

\toprule
Frequency & Name & Variant & Formula \\ 
\midrule
    27 & ${UCT}$ & 0 & $c \ \sqrt{\frac{2 \ ln  N(s)}{N(s,a)}}$ \\  \hline
    2 &  & 1 & $c  \ \sqrt{\frac{ln N(s)}{1 + N(s,a)}}$ \\ \hline
    1 &  & 2 & $c \ \frac{\sqrt{N(s)}}{1+N(s,a)}$ \\ \hline \hline
    39 & ${PUCT}$ & 0 & $c \ P(s,a) \frac{\sqrt{N(s)}}{1+N(s,a)}$ \\  \hline
    5 &  & 1 & $c \ P(s,a) \sqrt{\frac{2 \ ln N(s)}{N(s,a)}}$ \\ \hline
    3 &  & 2 & $c \ P(s,a) \frac{\sqrt{N(s)}}{N(s,a)}$  \\ \hline
    3 &  & 3 & $c \ P(s,a)^\mu \sqrt{\frac{N(s)}{1+N(s,a)}}$ \\ \hline
    2 &  & 4 & $c \ \frac{P(s,a)}{1 + N(s,a)}$ \\  \hline
    2 &  & 5 & $c \ P(s,a) \frac{\sqrt{N(s) + 1}}{N(s,a)+ 1}$  \\ \hline
    2 &  & 6 & $c \ P(s,a) \frac{N(s)}{1+N(s,a)}$ \\  \hline
    1 &  & 7 & $c \ P(s,a) \frac{\sqrt{N(s) + \epsilon}}{N(s,a)+ 1}$  \\ \hline
    1 &  & 8 & $c \ P(s,a) \sqrt{\frac{ln N(s) + 1}{1+N(s,a)}}$ \\  \hline
    1 &  & 9 & $c \ P(s,a) \sqrt{\frac{N(s)}{1+N(s,a)}}$ \\ \hline
    1 &  & 10 & $c \  P(s,a) \sqrt{\frac{ln N(s)}{1 + N(s,a)}}$ \\ \hline \hline
    2 & ${MuZero}$  & 0  & $P(s,a) \frac{\sqrt{N(s)}}{1+N(s,a)}\big[c_1 +ln \ \frac{N(s) + c_2 + 1}{c_2}\big]$ \\ \hline
    1 &  & 1 & $P(s,a) \frac{\sqrt{N(s)}}{1+N(s,a)}\big[c_1 +ln \ \frac{\sqrt{N(s) + c_2 + 1}}{c_2}\big]$ \\ \hline \hline 
    1 & ${UCT}_D$& 0 & $c \ A(s,a) \sqrt{\frac{2 \ ln \ N(s)}{N(s,a)}}$  \\ \hline \hline 
    1 & ${PUCT}_B$ & 0 & $c_1 \ P(s,a) \frac{\sqrt{N(s)}}{1+N(s,a)} + c_2 B(s, a)$ \\  

\botrule
\end{tabular}
\renewcommand{\arraystretch}{1}

\end{center}
\end{table*}
Among the surveyed publications, a large proportion still use (some variant of) the \gls{UCT} formula (see \cref{tab:uct_variations}), but \gls{PUCT} as it is used in AlphaZero \cite{silver_general_2018} (\gls{PUCT} variant 0 in \cref{tab:uct_variations}) is the most frequently used selection mechanism. There are a number of less frequently used \gls{PUCT} variations mostly concerning the presence of logarithms, constant factors in the numerator and denominator, and scope of the square root. These differences impact the overall magnitude of the exploration term as well as its decay as individual actions are visited more often (see \cref{fig:puct_uct}). It is difficult to judge the impact of different formulations on the search, since authors usually do not directly compare them. In a rare exception, Xu and Lieberherr \cite{xu_learning_2019} try both the AlphaZero \gls{PUCT} variant as well as PUCT variant 9 in \cref{tab:uct_variations} and report that the AlphaZero variant performs much better, although they do not quantify this difference. 

One notable \gls{PUCT} variant, variant 3, introduces a new constant $\mu$ which determines the impact of the prior probabilities as $P(s,a)^\mu$. This variant seems to have been independently suggested in \cite{shen_m-walk_2018,song_data_2020,wang_learning_2021}. 

Aside from \gls{UCT} and \gls{PUCT} variants, the MuZero \cite{schrittwieser_mastering_2020} selection formula or a variant of it is used by three authors. We further find two unique modifications of typical selection formulae that we cannot assign to any of the other groups: ${UCT}_D$ and ${PUCT}_B$. The former will be discussed at a later point. ${PUCT}_B$ aims to exploit the nature of deterministic single-player settings, in which future trajectories are not influenced by the choices of another player. In such cases, rather than simply looking at average state values, it may be advantageous to keep track of the best encountered values during the search. Making decisions based on average values is problematic because most of the actions in a given state may be bad choices, while one specific single action may be a good choice. On average, the value of the node will then be low, even though a promising child exists. In a deterministic setting, the best path can be executed reliably and, accordingly, it makes sense to choose nodes based on their expected best value rather than the average one. Deng et al. \cite{deng_neural-augmented_2022} design a selection formula that makes use of this fact, which we refer to as ${PUCT}_B$ here. In ${PUCT}_B$, the best value of an action is simply scaled by a constant and then added to \gls{PUCT} variant 0.

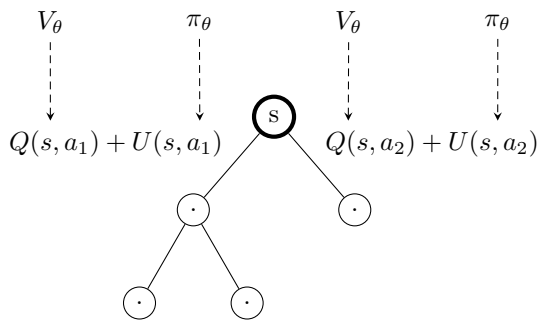
\begin{figure}
\centering
\resizebox{\columnwidth}{!}{
\input{Fig6}
}
\caption{Neural Selection. Each of the children of the current state $s$ are considered and the one maximizing $Q(s,a) + U(s,a)$ is chosen. Both $Q(s,a)$ and $U(s,a)$ may be influenced by neural guidance in some way. }
\label{fig:neural_selection}
\end{figure}

\paragraph{Neural guidance in the tree policy}
Neural guidance may be used in both the exploitation and the exploration part of \cref{eq:tree_policy} (see \cref{fig:neural_selection}). When used in the exploitation part, neural guidance is typically used to estimate $Q(s,a)$. This does not change how the selection mechanism works, only how the corresponding value is determined. Since value estimation is a function of the evaluation step, this kind of neural guidance will be explored in \cref{subsec:neural_eval} and not discussed further at this point. 

\begin{figure}
    \centering
    \includegraphics[width=0.9\columnwidth]{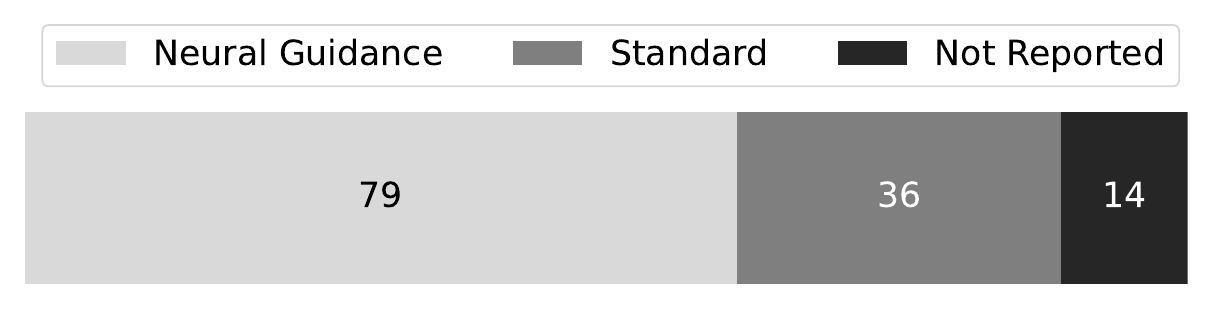}
    \caption{Proportion of choices in the selection phase among the surveyed articles. \emph{Standard} refers to some selection strategy that does not involve the use of learned functions.}
    \label{fig:neural_selection_freq}
\end{figure}

In the exploration part of \cref{eq:tree_policy}, neural guidance is typically used to determine the prior probabilities $P(s,a)$ in \gls{PUCT}-style formulae. As shown in \cref{fig:neural_selection_freq}, about 62\% of all reviewed articles report guiding the tree search in this way, with less than 30\% reporting selection phases without neural guidance. The remaining articles do not report how the selection phase is performed at all. While the latter can probably be interpreted as selection without neural guidance, we try to refrain from interpretations as much as possible and hence give separate categories for standard selection and unreported selection.  

Most approaches for neurally guided selection phases take the form described above, with the exception of a few special cases.  
Zombori et al. \cite{zombori_role_2021} argue that a learned policy network tends to make predictions with high confidence even if they are of low quality, which leads to a strong unfounded bias in the search. It may be more desirable to have a policy which makes less confident predictions if the prediction quality is not sufficiently high. To achieve this, they use maximum entropy reinforcement learning and use the resulting policy to compute prior probabilities for the selection phase.

In one exception, neural guidance occurs in a form other than providing prior probabilities, as can be seen in the ${UCT}_D$ formula in \cref{tab:uct_variations}. It is named after its use of a dueling network, which  produces action advantage estimates $A(s,a)$ in addition to state values. In ${UCT}_D$, the action advantages are used in place of the prior probabilities $P(s,a)$. Vaguely related to the reasoning of Zombori et al., the authors argue that a policy network trained with policy gradient methods tends to concentrate on the best action for a given state, while not assigning probabilities proportional to the expected usefulness of the other actions \cite{wang_task-completion_2020}. In other words, an overly low entropy policy vector may bias the search to an undesirable degree. In contrast, the action advantages do not overly focus on the best action. 

\subsection{Guided Expansion}

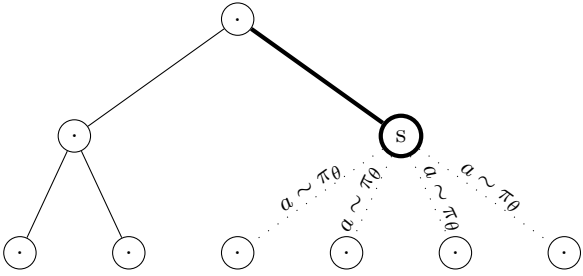
\begin{figure}
\centering
\input{Fig8}
\caption{Neural Expansion. When a leaf node is encountered, possible actions in the leaf node's state are sampled by some mechanism involving the learned policy. For every sampled action, a new child is created.}
\end{figure}

Once a leaf node $s_L$ is encountered during the selection phase in \gls{MCTS}, the expansion step is performed to create child nodes of $s_L$. Guidance by neural networks can be employed in this step as well to bias and hence speed-up the search. To avoid confusion, we will first give some details on possible alternate ways to implement the expansion step and only then return to the topic of neural guidance. 

As discussed in \cref{subsec:mcts}, nodes are typically expanded either one at a time whenever an expandable node $s_E$ is encountered, or all children of a node are expanded simultaneously if a leaf node $s_L$ is encountered.

Clearly, implementing an \gls{MCTS} approach requires deciding how many children are expanded at a given time. There is, however, an additional, related decision to be made: How many and which children are \emph{considered} for expansion? In the naive case, the search is free to choose any action from the set of all possible actions $A(s_L)$ in state $s_L$. However, it is also possible to be more selective in the expansion step. To limit the growth of the tree, only a limited number of children may be considered for expansion either randomly or according to some rule or heuristic. In other words, the search may be restricted to only choose actions from a set $\Tilde{A}(s_L) \subset A(s_L)$. This is especially relevant for continuous cases, where the number of potential children is infinite and necessarily has to be limited in some way. Once such a set $\Tilde{A}(s_L)$ has been defined, the corresponding child nodes may be expanded all at once when the leaf is encountered or one by one, whenever the expendable node is encountered during the tree search. 

Neural guidance during the expansion step is possible in both paradigms, i.e. when expanding on encountering a true leaf node and when expanding on encountering an expandable node. In the former case, neural guidance means using a learned policy to determine $\Tilde{A}(s_L)$, while in the latter case, neural guidance means choosing an action in $A(s_E)$ to create a new child node. Theoretically, it is possible to combine both of these approaches by first determining and saving $\Tilde{A}(s_L)$ when a leaf is encountered for the first time, but not expanding all corresponding nodes at this point. The children can then be expanded one by one whenever the node is encountered again by choosing some action $a \in \Tilde{A}(s_E)$. However, we do not observe this combined approach in the collected literature. 

\begin{figure}
    \centering
    \includegraphics[width=0.9\columnwidth]{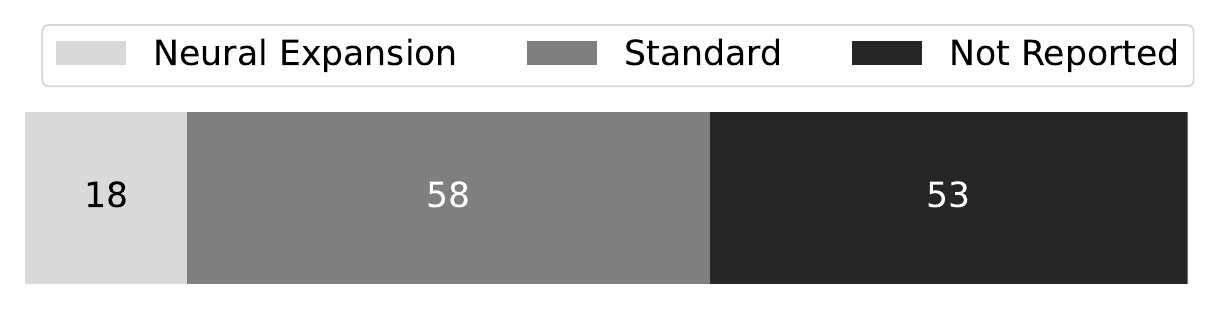}
    \caption{Proportion of choices in the expansion phase among the surveyed articles. \emph{Standard} refers to some form of expansion that does not involve the use of neural networks.}
    \label{fig:neural_expansion_freq}
\end{figure}

We do observe some form of neurally guided expansion in a sizeable portion of publications (see \cref{fig:neural_expansion_freq}) and categorize them in \cref{tab:neural_expansion}. In some additional examples, neurally guided expansion is used, but the specifics are not reported \cite{erikawa_mermaid_2021,genheden_aizynthfinder_2020,segler_planning_2018}. 

$\Tilde{A}(s_L)$ can be determined by randomly sampling actions from a learned policy, but it can also be determined by enumerating all actions in the policy distribution and choosing the top $k$ ones \cite{ishida_ai-driven_2022,segler_towards_2017,thakkar_datasets_2020}. In the approach of Thakkar et al. \cite{thakkar_datasets_2020}, the top $k$ actions with a cumulative policy probability of $0.995$ or at most $50$ actions are selected. In both types of neurally guided expansion, instead of a learned policy, a learned value function can of course be converted to a policy with a softmax operator, as is done in \cite{wang_scene_2020}.

\begin{table}[h]
\begin{center}
\caption{Types of neurally guided expansion.}
    \label{tab:neural_expansion}
\begin{tabular}{@{}p{0.4\columnwidth}p{0.45\columnwidth}@{}}

\toprule
Type & Publications \\ 
\midrule
Choosing $a \in A(s_E) $ & \cite{bai_hierarchical_2022}  \cite{dieb_optimization_2020}  \cite{hu_spear_2019} \cite{wang_scene_2020} \cite{yang_reinforcement_2020} \cite{lu_incorporating_2021} \\ \\
    Determining $\Tilde{A}(s_L)$ & \cite{fawzi_discovering_2022}  \cite{ishida_ai-driven_2022} \cite{lei_kb-tree_2021}  \cite{raina_learning_2022} \cite{riviere_neural_2021} \cite{segler_towards_2017} \cite{thakkar_datasets_2020} \cite{yang_chemts_2017} \cite{zhang_chemistry-informed_2022} \\ 

\botrule
\end{tabular}
\end{center}
\end{table}

While the exact impact of neurally guided expansion will vary from application to application, its general potential is demonstrated in \cite{riviere_neural_2021} who report that their computational time is 20 times reduced with neural expansion while achieving higher quality solutions.

\subsection{Guided Evaluation}
\label{subsec:neural_eval}

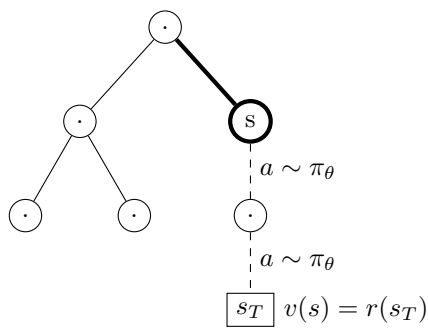
\begin{figure}
\centering
\input{Fig10}
\caption{Evaluation by learned policy roll-out: After arriving at leaf node with state $s$ according to the tree policy, the value of $s$ needs to be determined. Here, the learned policy $\pi_\theta$ is used to generate a roll-out by iteratively sampling actions until a terminal state $s_T$ is reached. The reward of this terminal state serves as an estimate for the value of $s$.}
\label{fig:neural_rollout}
\end{figure}

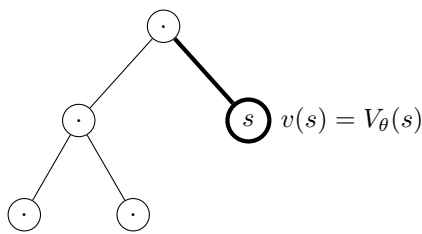
\begin{figure}
\centering
\input{Fig11}

\caption{Evaluation by learned value function: After arriving at leaf node with state $s$ by following the tree policy, the value of $s$ needs to be determined. Here, a learned value function $V_\theta$ is used to estimate the value of $s$ directly, without any need for a roll-out.}
\label{fig:neural_value_evaluation}
\end{figure}

The evaluation step in \gls{MCTS} serves to estimate the (state-)value of a leaf node encountered during the tree search. While it is often also called the roll-out step or the simulation step, its purpose is the value estimation of a leaf. Roll-outs or simulations are simply approaches to produce a value estimate. Here, we use the term evaluation, because not all evaluation approaches in the neural \gls{MCTS} literature are based on roll-outs. 

There are two obvious ways to use learned policies and value functions during the evaluation phase: a roll-out using the learned policy and a direct prediction by a learned value function. In the former, actions are iteratively sampled from the learned policy starting from the encountered leaf node until a terminal node is reached (see \cref{fig:neural_rollout}), while in the latter, the value of the leaf node is simply predicted by the learned value function without any roll-out (see \cref{fig:neural_value_evaluation}). 

\begin{figure}
    \centering
    \includegraphics[width=1.0\columnwidth]{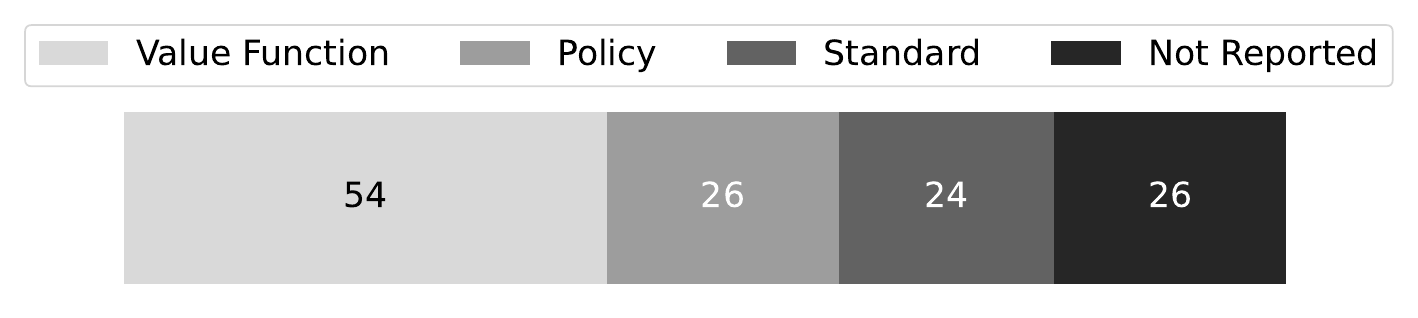}
    \caption{Proportion of choices in the evaluation phase among the surveyed articles. \emph{Value function} refers to an evaluation by a learned value function, \emph{policy} to a roll-out using the learned policy and \emph{standard} to a random roll-out.}
    \label{fig:neural_eval_freqs}
\end{figure}

Most authors use either learned policy or value functions as described above, with 54 occurrences of learned value functions and 26 occurrences of learned policy functions (see \cref{fig:neural_eval_freqs}) The remaining publications either do not use neural guidance for the evaluation phase or their approach is unclear. 

Among the neural evaluation approaches, some authors employ different evaluation approaches depending on how far the training has progressed. Song et al. \cite{song_data_2020} combine both approaches by performing roll-outs according to a learned policy network in the early phases of training and use a learned value network for estimation in later stages. Zhang et al. \cite{zhang_reinforcement_2020} use an initial phase of random roll-outs to pre-train a policy network and employ the learned policy network for roll-outs in later stages. 

In some cases, roll-outs are not performed by naively using a learned policy, but more complex roll-out procedures are still guided by learned functions. He et al. \cite{he_neurally-guided_2018} use a value network to guide a problem specific roll-out procedure, while Xing et al. \cite{xing_graph_2020} use a learned policy function to guide a beam search. Kumar et al. \cite{kumar_production_2021} combine a value estimate as predicted by a neural network with a domain-specific roll-out policy, motivated by the fact that their reward function is more fine-grained than those typically observed in board games. Finally, Lu et al. \cite{lu_incorporating_2021} use a value network in a symbolic regression task to estimate whether a leaf node merits further refinement by an optimization method, but their approach is highly problem-specific. 

Deng et al. \cite{deng_neural-augmented_2022} perform different evaluation approaches depending on the depth of the node to be evaluated. If the node is closer to the root of the tree, they use a neural network to estimate the value of the state, while they perform a random roll-out for nodes at deeper levels of the tree. Their experiments show that this hybrid approach can balance solution quality and computation time. Since neural inferences are associated with non-negligible computational cost, replacing them with random roll-outs can decrease the search time especially at deeper levels, where roll-outs will have short lengths. 

In the application described by He et al. \cite{he_circuit_2022}, episodes can result in either a successful or a failed solution. Successful solutions can still differ in quality and are rewarded accordingly. They perform a roll-out by a learned policy, but if this  results in a failed terminal state, they backtrack until they find a successful solution. They argue that this leads to better search efficiency because preceding trajectories are not repeated unnecessarily. 

Design choices in other \gls{MCTS} phases can also influence what is required from the evaluation phase. Kovari et al. \cite{kovari_design_2020} replace the exploitation term, i.e. the estimated value, of a UCT-style formula with the probability of taking an action as predicted by the policy network. Instead of a roll-out or direct value prediction, they hence simply predict this probability in the evaluation step.

\subsection{Guidance in Multiple Phases}

\begin{figure*}
    \centering
    \includegraphics[width=0.8\textwidth]{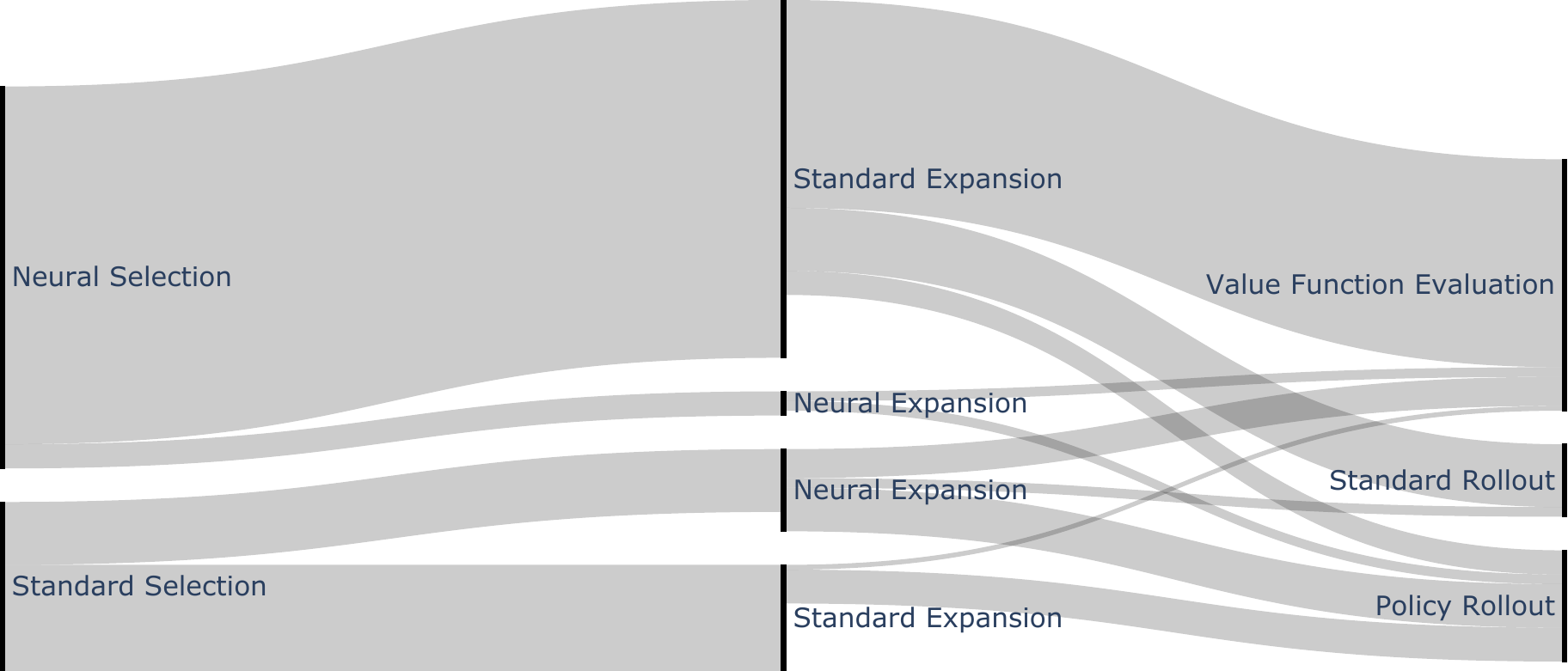}
    \caption{Parallel sets diagram of surveyed neural guidance configurations. Each vertical bar signifies an option in one of the \gls{MCTS} phases: selection (left), expansion (middle), evaluation (right). In the expansion step, each option (standard and neural expansion) is displayed twice to allow for easier tracing of the visualized configuration.}
    \label{fig:guidance_combinations}
\end{figure*}
As discussed above, neural guidance can be employed in the selection, expansion, and evaluation of phases of \gls{MCTS}. Of course, it is not necessary to limit this guidance to one phase at a time and different types of neural guidance can be combined in one approach. 

To gain an overview of how different types of neural guidance are typically combined, we visualize their use in \cref{fig:guidance_combinations}.

The most common approach is to guide the selection step, perform standard \gls{MCTS} expansion, and then use a learned value function for evaluation. Many other combinations exist, but none of them are used as often. When standard selection is used, the relative incidence of neural expansion is higher than when using neural selection. This could, however, be explained by the fact that the respective authors simply wanted to highlight the effect of neural expansion since it is often the main focus of their respective publications. 

Infrequently occurring combinations may indicate a need for further research. 

\subsection{Use of Dynamics Models}
Neural \gls{MCTS} is a model-based reinforcement learning approach, i.e., it requires access to a dynamics model of the environment to perform planning. Typically this dynamics model is given \cite{silver_mastering_2016,silver_mastering_2017} and can be readily used in the tree search. In contrast to a regular reinforcement learning environment as defined in, e.g. the OpenAI Gym standard \cite{brockman_openai_2016}, a dynamics model allows for the computation of the next state and reward given an arbitrary initial state and action to be executed. An environment, on the other hand, is in a specific state at any given time which can be influenced by actions, but does not allow for dynamics computations on arbitrary states. In other words, an environment is stateful, while a dynamics model is not. 

The difficulty of developing such a dynamics model will differ from application to application. In any case, its development will require some additional effort. To circumvent this, it is also possible to \emph{learn} the dynamics model and then use this learned model for planning in \gls{MCTS} \cite{schrittwieser_mastering_2020,van_eyck_guided_2013}. While this adds additional complexity to the training process, it can be helpful in scenarios where an exact and efficient model of the environment cannot be easily obtained. 

During our review we found that the vast majority of approaches utilize an existing dynamics model, but learned models also find some application in practice. 

For instance, Chen et al. \cite{chen_driving_2020} investigate an autonomous driving task where a model is needed to predict the vehicle state. In this case, the vehicle state is an image, meaning that the model needs to produce an output image given an input image (corresponding to the initial state) and an action. While such a model is not trivial to implement manually, Chen et al. \cite{chen_driving_2020} are able to train a convolutional neural network to serve this purpose. 

Similarly, Challita et al. \cite{challita_deep_2021} apply neural \gls{MCTS} to enable dynamic spectrum sharing between LTE and NR systems and report that this requires a model of individual schedulers for LTE and NR, which is not trivial to design. Instead, they learn the model in an approach similar to the one proposed in MuZero \cite{schrittwieser_mastering_2020}. That is, the dynamics are not computed on the raw observations, but on hidden representations, which are computed from the observations by a learned representation function.  This approach is also taken by others \cite{gabirondo-lopez_towards_2021,shuai_online_2021}, but dynamics models which work directly on the observations can also be observed \cite{dai_aoi-minimal_2022,wang_task-completion_2020}. 

In many cases, the dynamics model is not learned during neural \gls{MCTS} training, but trained separately in advance and then simply used for inference during the tree search \cite{carfora_dialogue_2020,fong_model-based_2021,ha_vehicle_2020,wang_meta-modeling_2019} 

As described above, one motivation to learn a dynamics model may be the difficulty of creating one manually. Another motivation is the speed with which a learned model can be evaluated \cite{sv_multi-objective_2022}. 

In some cases, the state transitions can be modelled fairly easily, while the computation of the reward is time-consuming. Some authors do not train a full dynamics model, but a scoring model, which can be used to assess the quality of a given solution quickly. In contrast to a learned value function, which can be used to evaluate newly expanded nodes at arbitrary depths, a scoring model only assigns a score to full solutions, i.e. terminal nodes. The resulting scores can then be used as training targets for the value function \cite{he_text_2018,srinivasan_artificial_2021,zhang_alphajoin_2020}.

\subsection{MCTS modifications}
We now turn our attention to selected modifications of typical (neural) \gls{MCTS} procedures as encountered during the review. 

\paragraph{Average and best values}
As briefly mentioned in \cref{subsec:guided_selection}, deterministic single-player settings pose different requirements than combinatorial games. Action selection based on the average value of a node will lead to sub-optimal results, because a strong child node can be surrounded by weak siblings. While the ${PUCT}_B$ mechanism described in \cref{subsec:guided_selection} is one option to address this, other authors have identified this issue as well and proposed their own solutions. 

Deng et al. \cite{deng_neural-augmented_2022} report that their final search results are usually worse than the best solutions found during roll-outs in empirical experiments. They point out that the final action selection after tree search is performed based on the node visit counts $N(s)$. To rectify the problem, they introduce an oversampling mechanism for good solutions. Whenever a solution is found which outperforms all previously found solutions during a roll-out, this solution will be given preference in subsequent selection phases for a certain amount of time, and will hence be visited more often.   

A simpler approach is taken by Peng et al. \cite{peng_lore_2022} and Xing et al. \cite{xing_solve_2020} to address the same problem. Here, the exploitation part of \cref{eq:tree_policy}, i.e. the average value of the node, is simply replaced with the best observed value for the node.  Fawzi et al. \cite{fawzi_discovering_2022} follow a similar strategy. 

Zhang et al. \cite{zhang_reinforcement-learning-based_2022} simply keep track of the maximum reward encountered in the search and the action sequence that lead to it, which is then returned after the search.  

\paragraph{Value normalization}
While combinatorial games lend themselves well to reward function formulations in the range $[-1, 1]$, in other applications, rewards are often less regular and sometimes completely unbounded. As mentioned earlier, the exploration constant $c$ needs to be tuned for different reward ranges \cite{browne_survey_2012}. Further, even with a perfectly tuned $c$, rewards outside of ranges like $[-1, 1]$ or $[0, 1]$ are typically not conducive to algorithm convergence \cite{shuai_post-storm_2023}. A number of authors therefore suggest to normalize Q-values according to the minimum and maximum values observed in the tree search until the current point \cite{qian_alphadrug_2022,shuai_post-storm_2023,xing_solve_2020}. 

\paragraph{The cost of neural inference}
The main idea behind neural \gls{MCTS} approaches is to increase the efficiency of the tree search with neural inferences. While neural inferences are not too computationally expensive individually, when performed in large numbers, the required computational time can add up to significant amounts. 

Deng et al. \cite{deng_neural-augmented_2022} vary the amount of neural inferences by switching neural guidance off during the search some proportion of the time. They find that neural guidance generally helps the search, but that further neural guidance after a certain point only increases computational cost without providing additional benefits. 

Designing mechanisms to limit the application of neural guidance to where it provides maximum benefits in a targeted way may be an interesting line of research. 

\subsection{MCTS Hyper-parameters}
\begin{figure}
    \centering
    \includegraphics[width=\columnwidth]{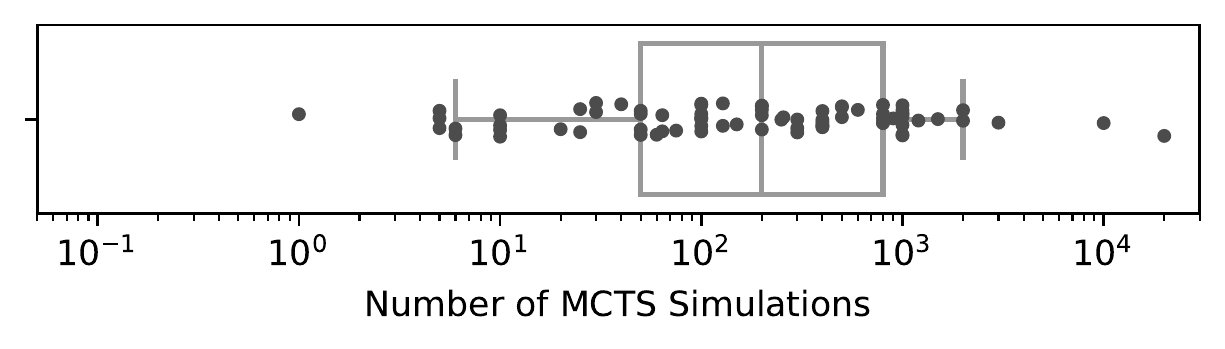}
    \includegraphics[width=\columnwidth]{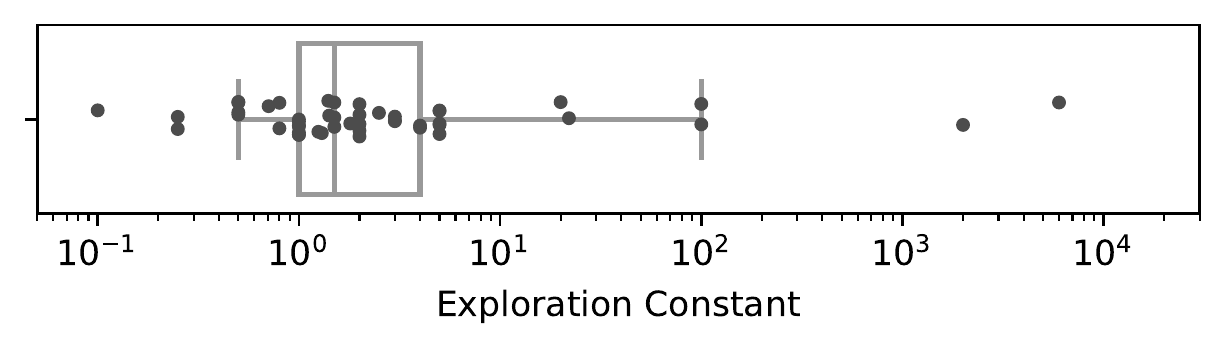}

    \caption{Distribution of reported \gls{MCTS} hyper-parameters: Number of \gls{MCTS} simulations (top) and exploration constant $c$ (bottom). Note that a logarithmic axis is used in both cases. Variation on the vertical axis is not meaningful, but contains random jitter for better visibility of individual data points.}
    \label{fig:mcts_hyper}
\end{figure}

One last aspect in the design of neural \gls{MCTS} approaches is the choice of appropriate hyper-parameters. While choosing hyper-parameters is highly problem-specific, it can nevertheless be useful to look at average hyper-parameter values to serve as a starting point and determine reasonable bounds for a problem-specific hyper-parameter optimization. To facilitate this, we summarize the values for the \gls{MCTS}-specific hyper-parameters found during our review in \cref{fig:mcts_hyper}. A large amount of variation in values exists for both the exploration constant $c$ used in \gls{UCT}-style selection formulae and the \gls{MCTS}-budget $n_{MCTS}$, i.e. the number of simulations or play-outs performed during \gls{MCTS} for a given time-step. 

It is known that the optimal choice of $c$ depends on the scale of the encountered rewards \cite{browne_survey_2012}. While reward scales vary wildly and are not always reported among the literature we survey, the most commonly reported reward scale is in the interval $[-1, 1]$, for which six different values of $c$ ranging from $0.5$ to $5.0$ are chosen. In other words, the scale of the rewards does not appear to be the only criterion on which authors choose hyper-parameter values. 

Instead of having a fixed value, in some cases, the exploration constant $c$ and the \gls{MCTS} budget $n_{MCTS}$ are changed dynamically depending on circumstances. 

Sometimes the exploration constant $c$ is decreased as the training progresses \cite{chen_optimizing_2022} \cite{zombori_role_2021}, presumably because later training iterations profit more from perfecting the current policy instead of performing further exploration. Wang et al. \cite{wang_towards_2020} tune $c$ dynamically based on the currently observed maximum Q-value to balance exploration and exploitation as Q-value estimates evolve and report a significant improvement in the observed results. 

Some problem settings feature varying instance sizes, where a larger instance size is generally associated with higher difficulty. Zhong et al. \cite{zhong_learning_2019} increase $c$ as the problem size grows, presumably because more exploration is required to adequately cover the larger state space. For similar reasons, they and others  \cite{wang_learning_2021,xing_graph_2020} further choose larger $n_{MCTS}$ for larger instance sizes. Wang et al. \cite{wang_parallel_2020} even explicitly parameterize $n_{MCTS}$ by the problem size. 

In many problems, the size of the remaining search space decreases with increasing depth of the tree. Hu et al.  \cite{hu_spear_2019} argue that the search budget should depend on the depth of the node the search starts from and present a mechanism that decays $n_{MCTS}$ with increasing tree depth.  

Fawzi et al. \cite{fawzi_discovering_2022} report an increase in $n_{MCTS}$ after a certain amount of training steps, presumably because later training iterations can profit more from a higher search budget and a larger proportion of the overall training time budget should hence be allocated to those later iterations.  

Some authors reduce the number of \gls{MCTS} simulations at test time. For instance, Chen et al. \cite{chen_driving_2020} reduce $n_{MCTS}$ by a factor of ten at test time compared to the training phase. They further limit the search depth in both phases, but do so to a larger degree during test time. 

In some cases, \gls{MCTS} is only used to train a policy network, which is then applied without further tree search at test time \cite{carfora_dialogue_2020,challita_deep_2021,chen_deeppursuit_2021,dai_reinforcement_2021,zhong_learning_2019}. This can be due to the specific requirements of the application, i.e., some applications require fast inferences at test time that render the application of tree search infeasible, but can still profit from \gls{MCTS} at training time \cite{challita_deep_2021}. In some applications, applying \gls{MCTS} at test time after having used it for training simply does not improve performance to a significant degree \cite{dai_reinforcement_2021}.   


\section{Discussion \& Conclusion}
\label{sec:discussion_conclusion}
While focusing on usages of neural \gls{MCTS} outside of games, we investigated the diversity in applications, their characteristics, and the design of employed neural \gls{MCTS} approaches by performing a systematic literature review.

With regard to \textbf{research question 1} posed in the introduction, we find that neural \gls{MCTS} is applied in a wide variety of domains to solve different problems. While most problems exhibit similar characteristics such as discrete time and actions, finite horizon, and deterministic transitions, many authors also demonstrate that neural \gls{MCTS} can be applied to problems with differing properties. 

The applications encountered during the review usually have slightly different requirements and properties than combinatorial games. This does affect the way solutions are designed, an aspect investigated as part of \textbf{research question 2}. The concept of self-play, for instance, is generally not applicable to single-player problems. In some cases, single-player problems can be modelled as multi-player problems, but this is the exception rather than the rule. It is possible to employ a mechanism called self-competition, which replicates the way self-play generates rewards in a single-player setting. 

Many authors further point out that selecting actions based on \emph{average} node values is not ideal in single-player deterministic environments. Different mechanisms, all based around tracking maximal node value, can be employed to adjust the typical \gls{MCTS} mechanisms in this regard. 

Compared to the neural \gls{MCTS} ecosystem for games, as well as the traditional \gls{RL} ecosystem, the neural \gls{MCTS} landscape is almost completely devoid of standardized implementations and components. Instead, almost all implementations are entirely custom-made. In a few exceptions, domain-specific implementations are reused by others, but they can only be applied to a very narrow set of problems (see e.g. \cite{yang_chemts_2017}). While this is understandable for more fundamental research, where implementations are inherently in flux, we believe that standardized components could significantly speed up progress on the applied research side. 

One reason for the lack of standardized components may be that the design of neural \gls{MCTS} methods varies substantially, beginning with their training approaches, to their use of self-play related mechanisms, forms of neural guidance, and other modifications to traditional \gls{MCTS} setups. A widely applicable neural \gls{MCTS} framework would have to be highly configurable to accommodate different disciplines and problem settings. While this is a difficult task, the traditional \gls{RL} ecosystem demonstrates that standards \cite{brockman_openai_2016} and publicly available implementations of algorithms \cite{raffin_stable-baselines3_2021} accelerate progress, and that flexible frameworks suitable for research \cite{bitter_karolos_2022} can be designed.

In response to \textbf{research question 3}, it can be concluded that the forms of neural guidance used in the game literature are often used in other applications as well. The most common type of guidance consists of neural selection and evaluation by a learned value function, just as in AlphaZero \cite{silver_general_2018}. Other types, and combinations of neural guidance can be found in the literature as well. Given the amount of variation in different neural \gls{MCTS} systems, a central question for practitioners is in how to set up their own systems depending on the characteristics of their applications. Ideally, we would be able to map observed problem characteristics to observed neural \gls{MCTS} configurations in order to provide a guideline for others to use. While some problem characteristics, e.g. the discrete or continuous nature of the action space, can be determined fairly reliably when reviewing existing publications, others are not so easy to ascertain. The breadth and depth of the (full) tree, for instance, are rarely reported explicitly. In some cases, it may be possible to infer them, but in a review with a multitude of different disciplines, trying to do so reliably is difficult. 

Some insights can nevertheless be derived from the collected literature. In games and beyond, it is clear that neural guidance can help to increase the efficiency of the tree search, but can also incur computational cost without much additional benefit in some situations. When and how to employ neural guidance should hence be carefully weighed. During the evaluation phase, for instance, the right choice of mechanism depends on the depth of the overall tree as well as the depth of the node to be evaluated. If the length of a roll-out will be short, it may be preferable over an estimate by a learned value function with associated inference cost. At what exact depth one may be preferable over the other will depend on the size of the employed neural network, which will influence the inference cost, as well as the quality of its predictions. Competing with a high quality estimate of a learned value function may require multiple roll-outs, since individual roll-outs are high variance estimates of a node's value. A good initial estimate can help focus the search on promising regions of the solution space, while a bad estimate can lead the search astray.

In applications with large tree breadth, neural guidance may be especially helpful in the expansion phase, where it can be used to prevent certain paths in the search tree from consideration altogether. Of course, this comes at the risk of cutting off high-quality solutions. Here as well, the quality of neural network predictions determines whether such an approach is sensible. Especially in applications with very large tree breadth, or even continuous domains, however, a search may not even be feasible without limiting the solution space to some degree. 

Clearly, many questions remain unanswered. Additionally, a purely backward looking review tends to summarize what has been done in the past, rather than what should have been done. What is presented in this review is therefore primarily a map of existing approaches and less so a collection of prescriptive knowledge. The results gathered in this review can, however, serve as a foundation for further experimental studies. Explicitly designing experiments for multiple applications with different properties, in which the factors identified in this review are systematically controlled, can serve to create a more robust understanding of the design of neural \gls{MCTS} approaches. From such an understanding, prescriptive rules (of thumb) can be derived to aid practitioners in making appropriate design choices for applications with given properties. 


\section*{Acknowledgements}
Funded by the Deutsche Forschungsgemeinschaft (DFG, German Research Foundation) under Germany's Excellence Strategy – EXC-2023 Internet of Production – 390621612.

\bibliographystyle{acm}
\bibliography{01_bibliography}

\clearpage
\begin{appendices}

\section{Implementations}
\label{sec:app_implementations}

\begin{table}[h]
\caption{Publicly accessible implementations of neural \gls{MCTS} approaches.} \label{tab:appendix_implementations}%
\begin{tabular}{@{}rp{0.8\textwidth}@{}}
\toprule
Source & Link to Implementation \\ 
\midrule
\cite{bai_hierarchical_2022} & \url{https://github.com/baifanxxx/NPMO-Rearrangement} \\ 
\cite{erikawa_mermaid_2021} & \url{https://github.com/sekijima-lab/mermaid} \\
\cite{gabirondo-lopez_towards_2021} & \url{https://github.com/werner-duvaud/muzero-general} (muzero general) \\
\cite{gauthier_deep_2020} & \url{https://github.com/HOL-Theorem-Prover/HOL} \\ 
\cite{genheden_aizynthfinder_2020} & \url{https://github.com/MolecularAI/aizynthfinder} \\
\cite{gros_real-time_2020} & \url{https://mgit.cs.uni-saarland.de/timopgros/carmanufacturin} \\
\cite{riviere_neural_2021} & \url{https://github.com/bpriviere/decision_making} \\ 
\cite{shao_alphaseq_2020} & \url{https://github.com/lynshao/AlphaSeq} \\
\cite{srinivasan_artificial_2021} & see \cite{yang_chemts_2017} \\
\cite{sv_multi-objective_2022} & \url{https://github.com/NREL/rlmolecule} \\
\cite{todi_adapting_2021} & \url{https://userinterfaces.aalto.fi/adaptive/} \\ 
\cite{wang_scene_2020} & \url{https://github.com/HanqingWangAI/SceneMover} \\ 
\cite{weingertner_monte_2020} & \url{https://github.com/PhilippeW83440/MCTS-NNET} \\ 
\cite{xiangyu_machine_2020} & see \cite{yang_chemts_2017} \\ 
\cite{xing_graph_2020} & \url{https://github.com/CMACH508/2020-GNN-MCTS-TSP} \\
\cite{yang_chemts_2017}  & \url{https://github.com/tsudalab/ChemTS} \\
\cite{zhang_chemistry-informed_2022} & \url{https://github.com/zbc0315/synprepy} \\ 
\cite{zombori_prolog_2020} & \url{https://github.com/zsoltzombori/plcop prolog} \\
\cite{zombori_role_2021} & \url{https://github.com/zsoltzombori/plcop} \\

\cite{zou_reinforcement_2019} & \url{https://github.com/zoulixin93/FMCTS} \\ 

\botrule
\end{tabular}

\end{table}
\end{appendices}

\end{document}

%% file: Fig1.tex
\begin{tikzpicture}[every tree node/.style={draw,circle},
   level distance=1.25cm,sibling distance=1cm,
   edge from parent path={(\tikzparentnode) -- (\tikzchildnode)}]
\Tree
[.\node{$s_0$};
    \edge[ultra thick] node[auto=right] {$a_{01}$};
    [.\node[ultra thick]{$s_1$};
       \edge[loosely dotted] node[midway,left] {$a_{11}$};
       [.\node{$s_3$}; ]
       \edge[loosely dotted] node[midway,right] {$a_{12}$};
       [.\node{$s_4$}; ]
        ]
    \edge node[auto=left] {$a_{02}$};
    [.\node{$s_2$};
    \edge node[midway,left] {$a_{13}$};
       [.\node{$s_5$}; 
            \edge node[midway,left] {$a_{21}$};
           [.\node[rectangle]{$s_7$}; ]
           \edge node[midway,right] {$a_{22}$};
           [.\node[rectangle]{$s_8$}; ] 
       ]
       \edge node[midway,right] {$a_{14}$};
       [.\node{$s_6$};
           \edge node[midway,left] {$a_{23}$};
           [.\node[rectangle]{$s_9$}; ]
           \edge node[midway,right] {$a_{24}$};
           [.\node[rectangle]{$s_{10}$}; ]
       ]
    ]
]
\end{tikzpicture}

%% file: Fig2.tex
\begin{tikzpicture}[
    node distance=5mm and 5mm,
box/.style = {draw, minimum height=12mm, align=center},
sy+/.style = {yshift= 2mm}, 
sy-/.style = {yshift=-2mm},
every edge quotes/.style = {align=center}
                        ]
\node (n1) [box]             {\textbf{Keyword} \\ \textbf{Search}};                        
\node (n2) [box,below=of n1] {\textbf{Initial Set} \\ Web of Science: 473 \\ Scopus: 721 \\ IEEExplore: 249 \\ ScienceDirect: 42 \\ PubMed: 59};
\node (n3) [box,right=of n2, minimum height=1.7cm] {\textbf{De-} \\ \textbf{duplication} \\ 821 };
\node (n4) [box,right=of n3, minimum height=1.7cm] {\textbf{Abstract} \\ \textbf{Screening} \\ 311 };
\node (n5) [box,right=of n4, minimum height=1.7cm] {\textbf{Full-Text} \\ \textbf{Screening} \\ 117 };
\node (n6) [box,right=of n5, minimum height=1.7cm] {\textbf{Final Set} \\  129 };

\node (n7) [box,above=of n6, minimum height=1.7cm] {\textbf{Backward} \\ \textbf{Search} \\6 };
\node (n8) [box,below=of n6, minimum height=1.7cm] {\textbf{Forward} \\ \textbf{Search} \\6 };

\draw[thick,-Triangle] (n1.south) to [above,] (n2.north);
\draw[thick,-Triangle] ([sy+] n2.east) to [above,] ([sy+] n3.west);
\draw[thick,-Triangle] ([sy+] n3.east) to [above,] ([sy+] n4.west);
\draw[thick,-Triangle] ([sy+] n4.east) to [above,] ([sy+] n5.west);
\draw[thick,-Triangle] ([sy+] n5.east) to [above,] ([sy+] n6.west);
\draw[thick,-Triangle,to path={|- (\tikztotarget)}] (n5.north) to [above,] (n7.west);
\draw[thick,-Triangle,to path={|- (\tikztotarget)}] (n5.south) to [above,] (n8.west);

\draw[thick,-Triangle] (n7.south) to [above,] (n6.north);
\draw[thick,-Triangle] (n8.north) to [above,] (n6.south);

    \end{tikzpicture}

%% file: Fig3.tex
\tikzstyle{component} = [draw, text width=4em, fill=gray!10, text centered,
    minimum height=8em, rounded corners]
\def\arrdist{1.5em}

\begin{tikzpicture}
    \node (mcts) [component] {MCTS};
    \node (neural_net) [component, xshift=-10em] {$f_\theta$};
    
    \draw[->] ([yshift=\arrdist]neural_net.east) -- node [above] {$\pi_\theta$} ([yshift=\arrdist]mcts.west);
    \draw[->] ([yshift=-\arrdist]neural_net.east) -- node[below] {$V^\pi_\theta$} ([yshift=-\arrdist]mcts.west);

    \path [draw, ->] ([yshift=-\arrdist]mcts.east) -- node[below] (mcts_v) {$V^{\pi^*}_{MCTS}$} ([yshift=-\arrdist]6em, 0);
    \path [draw, ->] ([yshift=\arrdist]mcts.east) -- node[above] (mcts_pi) {$\pi^{*}_{MCTS}$} ([yshift=\arrdist]6em, 0);

    \path [draw, ->, densely dotted, rounded corners] (mcts_pi.north) |- (0, 5.5em) -| (neural_net.north);
    \path [draw, ->, densely dotted, rounded corners] (mcts_v.south) |- (0, -5.5em) -| (neural_net.south);
\end{tikzpicture}

%% file: Fig6.tex
\begin{tikzpicture}[every tree node/.style={draw,circle},
   level distance=1.25cm,sibling distance=1cm,
   edge from parent path={(\tikzparentnode) -- (\tikzchildnode)}]
\Tree
[.\node[ultra thick] {s};
    \edge node[auto=right] (PUCT1) {$Q(s, a_1) + U(s, a_1)$}; 
    [..
       \edge node[midway,left] {};
       [.. ]
       \edge node[midway,right] {};
       [.. ]
        ]
    \edge node[auto=left] (PUCT2) {$Q(s, a_2) + U(s, a_2)$};
    [.\node {.};     
    ]
]
\begin{scope}[yshift=0.5in]
\node (PR) at (3,0){$\pi_\theta$};
\node (VR) at (1,0){$V_\theta$};
\draw[-stealth, densely dashed] (PR.south) -- ([xshift=2.5em]PUCT2.north);
\draw[-stealth, densely dashed] (VR.south) -- ([xshift=-3.2em]PUCT2.north);

\node (PL) at (-3,0){$V_\theta$};
\node (VL) at (-1,0){$\pi_\theta$};
\draw[-stealth, densely dashed] (PL.south) -- ([xshift=-2.5em]PUCT1.north);
\draw[-stealth, densely dashed] (VL.south) -- ([xshift=3.2em]PUCT1.north);
\end{scope}
\end{tikzpicture}

%% file: Fig8.tex
\begin{tikzpicture}[every tree node/.style={draw,circle},
   level distance=1.55cm,sibling distance=1cm, label distance=5mm, 
   edge from parent path={(\tikzparentnode) -- (\tikzchildnode)}]
\Tree
[..
    \edge node[auto=right] {};
    [..
       \edge node[midway,left] {};
       [.. ]
       \edge node[midway,right] {};
       [.. ]
        ]
    \edge[ultra thick] node[auto=left] {};
    [.\node[ultra thick] {s};        
        \edge[loosely dotted] node[sloped, above=-1mm] {\small $a \sim \pi_\theta$};
        [.. ]    
        \edge[loosely dotted] node[xshift=-0.25em, sloped,  above=-1mm] {\small $a \sim \pi_\theta$};
        [.. ]
        \edge[loosely dotted] node[xshift=0.25em, sloped, above=-1mm] {\small $a \sim \pi_\theta$};
        [.. ]    
        \edge[loosely dotted] node[sloped, above=-1mm] {\small $a \sim \pi_\theta$};
        [.. ]
    ]
]
\end{tikzpicture}

%% file: Fig10.tex
\begin{tikzpicture}[every tree node/.style={draw,circle},
   level distance=1.25cm,sibling distance=1cm,
   edge from parent path={(\tikzparentnode) -- (\tikzchildnode)}]
\Tree
[..
    \edge node[auto=right] {};
    [..
       \edge node[midway,left] {};
       [.. ]
       \edge node[midway,right] {};
       [.. ]
        ]
    \edge[ultra thick] node[auto=left] {};
    [.\node[ultra thick] {s};        
        \edge[dashed] node[midway,right] {$a \sim \pi_\theta$};
        [.. 
            \edge[dashed] node[midway,right] {$a \sim \pi_\theta$};
            [.\node[draw, rectangle,label=right:{$v(s) = r(s_T)$}]{$s_T$}; ]
        ]        
    ]
]
\end{tikzpicture}

%% file: Fig11.tex
\begin{tikzpicture}[every tree node/.style={draw,circle},
   level distance=1.25cm,sibling distance=1cm,
   edge from parent path={(\tikzparentnode) -- (\tikzchildnode)}]
\Tree
[..
    \edge node[auto=right] {};
    [..
       \edge node[midway,left] {};
       [.. ]
       \edge node[midway,right] {};
       [.. ]
        ]
    \edge[ultra thick] node[auto=left] {};
    [.\node [ultra thick, label=right:{$v(s) = V_\theta(s)$}] {$s$};  ]
]
\end{tikzpicture}